\documentclass[10pt,twocolumn,letterpaper]{article}
	\usepackage{cvpr}
	\usepackage[title]{appendix}
	\usepackage{times}
	\usepackage{epsfig}
	\usepackage{graphicx}
	\graphicspath{ {Figures/}{Figures_select/}{Figures_video_select/} }
	\usepackage{amsmath}
	\usepackage{relsize}
	\usepackage{amssymb}
	\usepackage{tabularx,booktabs}
	\usepackage{multirow}
    \newcolumntype{L}{>{\centering\arraybackslash}X}
    \newcolumntype{C}{>{\centering\arraybackslash}X}

	\DeclareMathOperator*{\softmax}{softmax}

    \newcommand{\ignore}[1]{}
	
	\usepackage{xcolor}
	\definecolor{todo}{rgb}{1,.5,0} 
	
	\usepackage[pagebackref=true,breaklinks=true,letterpaper=true,colorlinks,bookmarks=false]{hyperref}
	\usepackage[british,UKenglish,USenglish,english,american]{babel}
	
	\cvprfinalcopy 
	
	\def\cvprPaperID{4128} 

	\ifcvprfinal\fi
	
\begin{document}

    \newenvironment{absolutelynopagebreak}
      {\par\nobreak\vfil\penalty0\vfilneg
       \vtop\bgroup}
      {\par\xdef\tpd{\the\prevdepth}\egroup
       \prevdepth=\tpd}
	
	\title{Deep Exemplar-based Video Colorization}
	\author{Bo Zhang$^1$
	\thanks{Author did this work during the internship at Microsoft Research Asia. Email: zhangboknight@gmail.com}
	, Mingming He$^{1,5}$, Jing Liao$^{2}$, Pedro V. Sander$^1$, Lu Yuan$^{3,4}$, Amine Bermak$^{1,6}$, Dong Chen$^3$ \\
	$^1$Hong Kong University of Science and Technology \quad
	$^2$City University of Hong Kong  \\
	$^3$Microsoft Research Asia \quad $^4$Microsoft AI Perception and Mixed Reality\\
	$^5$USC Institute for Creative Technologies \quad $^6$Hamad Bin Khalifa University
	}
	
	\maketitle
	\thispagestyle{empty}
	\begin{abstract}
	This paper presents the first end-to-end network for exemplar-based video colorization. The main challenge is to achieve temporal consistency while remaining faithful to the reference style. To address this issue, we introduce a recurrent framework that unifies the semantic correspondence and color propagation steps. Both steps allow a provided reference image to guide the colorization of every frame, thus reducing accumulated propagation errors. Video frames are colorized in sequence based on the colorization history, and its coherency is further enforced by the temporal consistency loss. All of these components, learned end-to-end, help produce realistic videos with good temporal stability. Experiments show our result is superior to the state-of-the-art methods both quantitatively and qualitatively.
	\end{abstract}
	
	\section{Introduction}
	Prior to the advent of automatic colorization algorithms, artists revived legacy images or videos through a careful manual process. Early image colorization methods relied on user-guided scribbles~\cite{levin2004colorization,yatziv2004fast,huang2005adaptive,qu2006manga,luan2007natural} or a sample reference~\cite{welsh2002transferring, bugeau2014variational,liu2008intrinsic, chia2011semantic, gupta2012image, charpiat2008automatic,ironi2005colorization,tai2005local} to address this ill-posed problem, and more recent deep-learning works~\cite{cheng2015deep,iizuka2016let,larsson2016learning, zhang2016colorful, deshpande2015learning, zhao2018pixel, baldassarre2017deep} directly predict colors by learning color-semantic relationships from a large database.

    A more challenging task is to colorize legacy videos. Independently applying image colorization (\eg, ~\cite{iizuka2016let,larsson2016learning,zhang2016colorful}) on each frame often leads to flickering and false discontinuities. Therefore there have been some attempts to impose temporal constraints on video colorization. A na\"{i}ve approach is to run a temporal filter on the per-frame colorization results during post-processing~\cite{bonneel2015blind,lai2018learning}, which can alleviate the flickering but cause color fading and blurring. Another set of approaches propagate the color scribbles across frames using optical flow~\cite{levin2004colorization,yatziv2004fast,sheng2014video,dougan2015key,paul2017spatiotemporal}. However, scribbles propagation may be not perfect due to flow error, which will induce some visual artifacts. The most recent methods assume that the first frame is colorized and then propagate its colors to the following frames~\cite{jampani2017video,vondrick2018tracking,liu2018switchable,meyer2018deep}. This is effective to colorize a short video clip, but the errors will progressively accumulate when the video is long. These existing techniques are generally based on color propagation and do not consider the content of all frames when determining the colors.
    \ignore{
    They can be classified into three categories. The first type of methods is to run a temporal filter on the per-frame colorization results as a post-processing step~\cite{bonneel2015blind,lai2018learning}. This can refine the temporal flickering at the cost of color fading and blurriness. The second class propagates the color scribbles from one frame to the following according to optical flow ~\cite{levin2004colorization,yatziv2004fast,sheng2014video,dougan2015key,paul2017spatiotemporal}. However, scribbles drawn from one specific image may not be suitable for other frames, which will induce some visual artifacts. The most recent and popular way to colorize videos is to assume the first frame is colorized and then propagate its colors to the following frames in sequence ~\cite{jampani2017video,vondrick2018tracking,liu2018switchable,meyer2018deep}. This is effective to colorize a short video, but the errors will progressively accumulate when the video is long.
    }
    
We instead propose a method to colorize video frames jointly considering three aspects, instead of solely relying on the previous frame. First, our method takes the result of the previous frame as input to preserve temporal consistency. Second, our method performs colorization using an exemplar, allowing a provided reference image to guide the colorization of every frame and reduce accumulation error. Thus, finding semantic correspondence between the reference and every frame is essential to our method. Finally, our method leverages large-scale data from learning, so that it can predict natural colors based on the semantics of the input grayscale image when no proper matching is available in either the reference image or the previous frame.

To achieve the above objectives, we present the first end-to-end convolutional network for exemplar-based video colorization. It is a recurrent structure that allows history information for maintaining temporal consistency. Each state consists of two major modules: a correspondence subnet to align the reference to the input frame based on dense semantic correspondences, and a colorization subnet to colorize a frame guided by both the colorized result of its previous frame and the aligned reference. All subnets are jointly trained, yielding multiple benefits. First, the jointly trained correspondence subnet is tailored for the colorization task, thus achieving higher quality. Second, it is two orders of magnitude faster than the state-of-the-art exemplar-based colorization method~\cite{he2018deep} where the reference is aligned in a pre-processing step using a slow iterative optimization algorithm~\cite{liao2017visual}. Moreover, the joint training allows adding temporal constraints on the alignment as well, which is essential to consistent video colorization. This entire network is trained with novel loss functions considering natural occurrence of colors, faithfulness to the reference, spatial smoothness and temporal coherence.
    
    The experiments demonstrate that our video colorization network outperforms existing methods quantitatively and qualitatively. Moreover, our video colorization allows two modes. If the reference is a colorized frame in the video, our network will perform the same function as previous color propagation methods but in a more robust way. More importantly, our network supports colorizing a video with a color reference of a different scene. This allows the user to achieve customizable multimodal results by simply feeding various references, which cannot be accomplished in previous video colorization methods. 
    \ignore{
    In summary, our contributions are as follows:
    \begin{itemize}
        \item We propose the first end-to-end convolutions network for exemplar-based image and video colorization. This lets our method to achieve better quality while running with hundreds of times faster compared to previous exemplar-based colorization methods.
        \item Our video colorization results are temporally consistent even for long video sequences because our network colorizes a frame based on both the previous result and the reference.
         \item A novel loss function considering naturalness, faithfulness to the reference, spatial smoothness and temporal consistence is proposed to train exemplar-based colorization.
        \item Our colorization allows the user to customize results by simply feeding different references.
	\end{itemize}
	}
	
	\section{Related work}
	\paragraph{Interactive Colorization.}
    Early colorization methods focus on using local user hints in the form of color points or strokes~\cite{levin2004colorization,yatziv2004fast,huang2005adaptive,qu2006manga,luan2007natural}. The local color hints are propagated to the entire image according to the assumption that coherent neighborhoods should have similar colors. These pioneering works rely on the hand-crafted low-level features for the color propagation. Recently, Zhang and Zhu et al.~\cite{zhang2017real} proposed to employ deep neural networks to propagate the user edits by incorporating semantic information and achieve remarkable quality. However, all of these user-guided methods require significant manual interactions and aesthetic skills to generate plausible colorful images, making them unsuitable for colorizing images massively.\vspace{-0.7em}
    
    \paragraph{Exemplar-based Colorization.}
    Another category of work colorizes the grayscale images by transferring the color from the reference image in a similar content. The pioneering work~\cite{welsh2002transferring} transfers the chromatic information to the corresponding regions by matching the luminance and texture. In order to achieve a more accurate local transfer, various correspondence techniques have been proposed by matching low-level hand-crafted features~\cite{bugeau2014variational, liu2008intrinsic, chia2011semantic, gupta2012image, charpiat2008automatic, ironi2005colorization, tai2005local}. Still, these correspondence methods are not robust to complex appearance variations of the same object because low-level features do not capture semantic information. More recent works~\cite{he2017neural,he2018deep} rely on the Deep Analogy method~\cite{liao2017visual} to establish the semantic correspondence and then refine the colorization by solving Markov random field model~\cite{he2017neural} or a neural network~\cite{he2018deep}. In those works, the correspondence and the color propagation are optimized independently, therefore visual artifacts tend to arise due to correspondence error. On the contrary, we unify the two stages within one network, which is trained end-to-end and produces more coherent colorization results. 
    \vspace{-0.7em}
    
    \paragraph{Fully Automatic Colorization.}
    With the advent of deep learning techniques, various fully automatic colorization methods have been proposed to learn a parametric mapping from grayscale to color using large datasets~\cite{cheng2015deep,iizuka2016let,larsson2016learning, zhang2016colorful, deshpande2015learning, zhao2018pixel, baldassarre2017deep}. These methods predict the color by incorporating the low and high-level cues and have shown compelling results. However, these methods lack the modelling of color ambiguity and thus cannot generate multimodal results. In order to address these issues, diverse colorization methods have been proposed using the generative models~\cite{isola2017image,deshpande2017learning,messaoud2018structural,guadarrama2017pixcolor,royer2017probabilistic}. However, all of these automatic methods are prone to produce visual artifacts such as color bleeding and color washout, and the quality may significantly deteriorate when colorizing objects out of the scope of the training data. \vspace{-0.7em}
    
    \paragraph{Video Colorization.}
    Comparatively, there has been much less research effort focused on video colorization. Existing video colorization can be classified into three categories. The first is to post-process the framewise colorization with a general temporal filter~\cite{bonneel2015blind,lai2018learning}, but these works tend to wash out the colors. Another class of methods propagate the color scribbles to other frames by explicitly calculating the optical flow~\cite{levin2004colorization,yatziv2004fast,sheng2014video,dougan2015key,paul2017spatiotemporal}. However, scribbles drawn from one specific image may not be suitable for other frames. Another category of video colorization methods use one colored frame as an example and colorize the following frames in sequence. While conventional methods rely on hand-crafted low-level features to find the temporal correspondence~\cite{jacob2009colorization,ben2015approximate,xia2016robust}, a recent trend is to use a deep neural network to learn the temporal propagation in a data-driven manner~\cite{jampani2017video,vondrick2018tracking,liu2018switchable,meyer2018deep}. These approaches generally achieve better quality. However, a common issue of these video color propagation methods is that the color propagation will be problematic if it fails on a particular frame. Moreover, these methods require a good colored frame to bootstrap, which can be challenging in some scenes, particularly when it is dynamic and with significant variations. By contrast, our work refers to an example reference image during the entire process, thus not relying solely on color propagation from previous frames. It therefore yields more robust results, particularly for longer video clips.
	
	\section{Method}
	\subsection{Overall framework}
	We denote the grayscale video frame at time $t$ as  $x^l_t \in \mathbb{R}^{H\times W \times 1}$, and the reference image as $y^{lab} \in \mathbb{R}^{H\times W \times 3}$. Here, $l$ and $ab$ represent the luminance and chrominance in LAB color space, respectively. In order to generate temporally consistent videos, we let the network, denoted by $\mathcal{G}_{V}$, colorize video frames based on the history. Formally, we formulate the colorization for the frame $\tilde{x}_{t}^{l}$ to be conditional on both the colorized last frame $\tilde{x}_{t-1}^{lab}$ and the reference $y^{lab}$:
    \begin{equation}
    \begin{split}
    \tilde{x}_{t}^{ab} & = \mathcal{G}_{V}(x_t^l|\tilde{x}_{t-1}^{lab},y^{lab})
    \end{split}
    \label{eq:video_colorization}
	\end{equation}
    
\begin{figure}[!tb]
    \centering
    \includegraphics[width=0.75\linewidth]{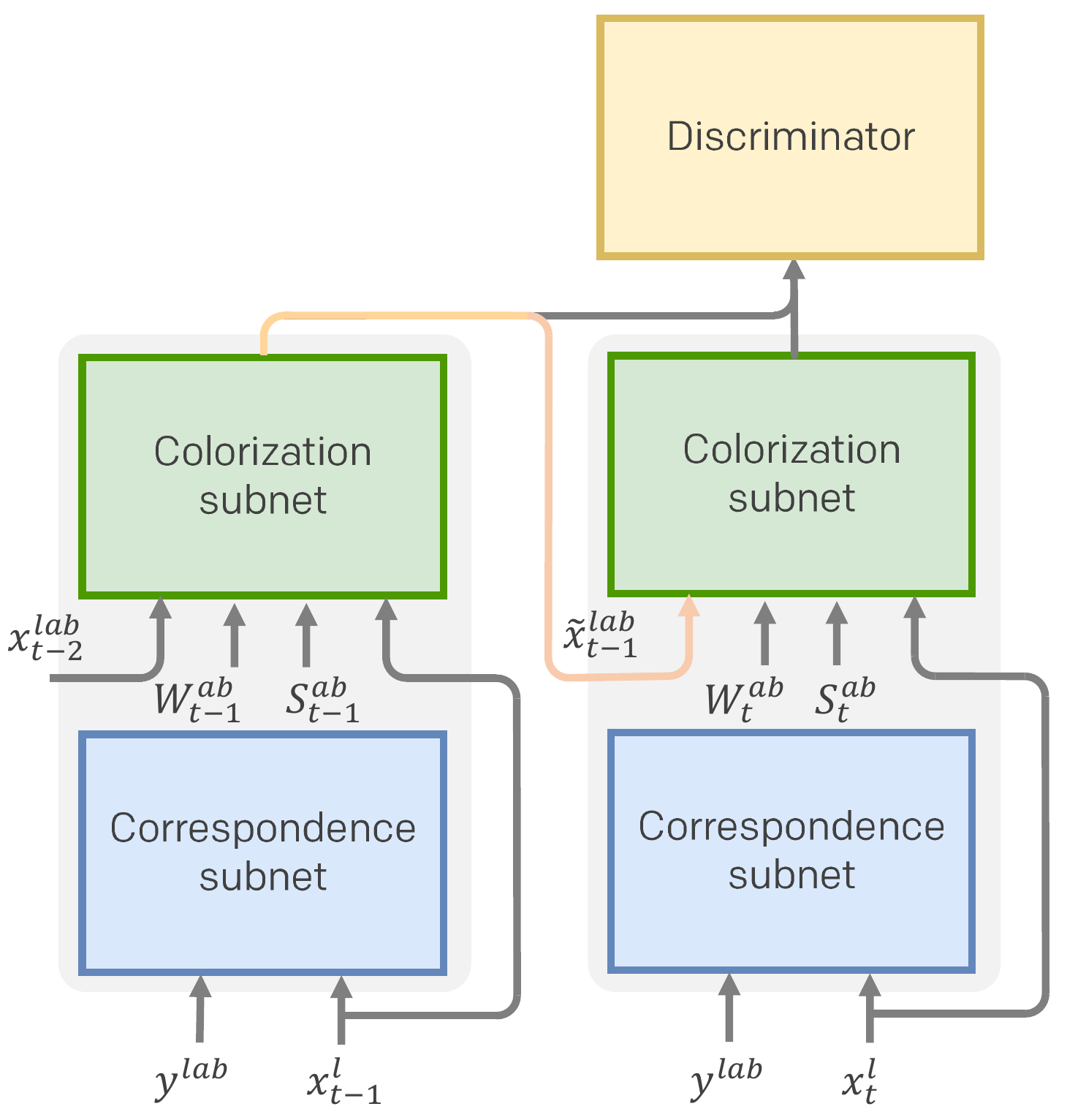}
    \caption{The framework of our video colorization network. The network consists of two subnets: correspondence subnet and colorization subnet. The colorization for the frame $x_t^l$ is conditional on the previous colorized frame $x_{t-1}^l$}.\vspace{-1.5em}
    \label{fig:network1}
    \end{figure}

    The pipeline for video colorization is shown in Figure~\ref{fig:network1}. We propose a two-stage network which consists of two subnets -  correspondence network $\mathcal{N}$ and colorization network $\mathcal{C}$. At time $t$, first $\mathcal{N}$ aligns the reference color $y^{ab}$ to $x^l_t$ based on their semantic correspondences, and yields two intermediate outputs: the warped color $\mathcal{W}^{ab}$ and a confidence map $\mathcal{S}$ measuring the correspondence reliability. Then $\mathcal{C}$  uses the warped intermediate results along with the colorized last frame $\tilde{x}_{t-1}^{lab}$ to colorize $\tilde{x}_{t}^{l}$. Thus, the network colorizes the video frames in sequence and Eq.~\ref{eq:video_colorization} can be expressed as:
\begin{equation}
\begin{split}
\tilde{x}_{t}^{ab} & = \mathcal{C}(x_t^l, \mathcal{N}(x_t^l, y^{lab})|\tilde{x}_{t-1}^{lab})
\end{split}
\end{equation}
    

	\begin{figure*}[!tb]
	\centering
	\includegraphics[width=1.0\linewidth]{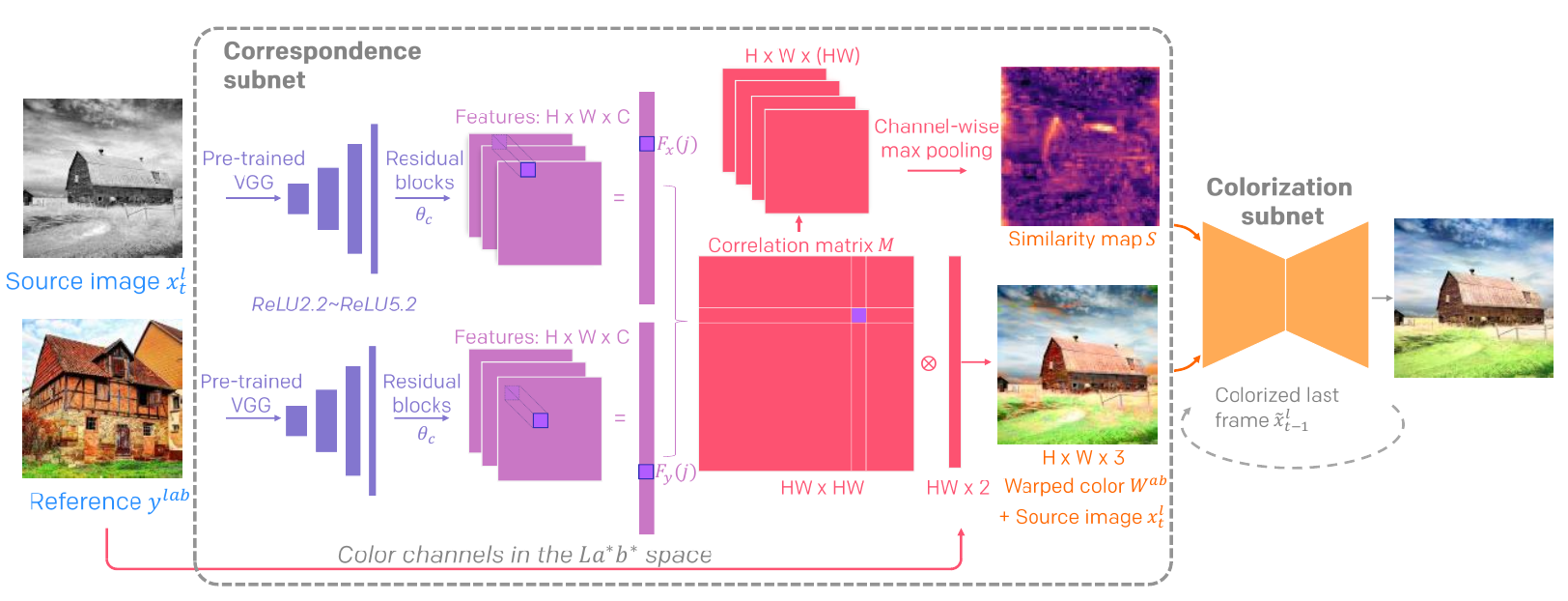}
	\caption{The detailed diagram of the proposed network. The correspondence subnet finds the correspondence of source image $x_t^{l}$ and reference image $y^{lab}$ in the deep feature domain, and aligns the reference color accordingly. Based on the intermediate result of the correspondence map along with the last colorized frame, the colorization subnet predicts the color for the current frame. }
	\label{fig:network}
	\end{figure*}

\subsection{Network architecture}
\label{Network architecture}
Figure~\ref{fig:network} illustrates the two-stage network architecture. Next we describe these two sub networks.\vspace{-0.5em}
	
\paragraph{Correspondence Subnet.} 
We build the semantic corresondence between $x^l_t$ and $y^{ab}$ using the deep features extracted from the VGG19~\cite{simonyan2014very} pretrained on image classification. In $\mathcal{N}$, we extract the feature maps from layers of $relu2\_2$, $relu3\_2$, $relu4\_2$ and $relu5\_2$ for both $x^l$ and $y^{ab}$. The multi-layer feature maps are concatenated to form features $\Phi_{x}, \Phi_{y} \in \mathbb{R}^{H\times W \times C} $ for $x^l_t, y^{ab}$ respectively. Features $\Phi_{x}$ and $\Phi_{y}$ are fed into several residual blocks to better exploit the features from different layers, and the outputs are reshaped into two feature vectors: $F_x, F_y \in \mathbb{R}^{HW \times C}$ for $x^l_t$ and $y^{ab}$ respectively. The residual blocks, parameterized by $\theta_{\mathcal{N}}$, share the same weights for $x^l_t$ and $y^{ab}$.

Given the feature representation, we can find dense correspondence by calculating the pairwise similarity between the features of $x^l_t$ and $y^{ab}$. Formally, we compute a correlation matrix $\mathcal{M}\in \mathbb{R}^{HW\times HW}$ whose elements characterize the similarity of $F_x$ at position $i$ and $F_y$ at $j$:
\begin{equation}
\mathcal{M}(i,j) = \frac{({F_x(i)}-\mu_{F_x})\cdot({F_y(j)}-\mu_{F_y})}{\Vert{F_x(i)}-\mu_{F_x} \Vert_2  \ \Vert{F_y(j)}-\mu_{F_y}\Vert_2}
\end{equation}
where $\mu_{F_x}$ and $\mu_{F_y}$ represent mean feature vectors. We empirically find such normalization makes the learning more stable. Then we can warp the reference color $y^{ab}$ towards $x^l_t$ according to the correlation matrix. We propose to calculate the weighted sum of $y^{ab}$ to approximate the color sampling from $y^{ab}$:
\begin{equation}
\mathcal{W}^{ab}(i) = \sum_j \softmax_j(\mathcal{M}(i,j)/ \tau) \cdot y^{ab}(j)
\label{eq:nonlocal}
\end{equation}
We set $\tau=0.01$ so that the row vector $\mathcal{M}(i,\cdot)$ approaches to one-hot vector and weighted color $\mathcal{W}^{ab}$ approximates selecting the pixel in the reference with largest similarity score. The resulting vector $\mathcal{W}^{ab}$ serves as an aligned color reference to guide the colorization in the next step. Note that Equation~\ref{eq:nonlocal} has a close relationship with the non-local operator proposed by Wang et al.~\cite{wang2017non}. The major difference is that the non-local operator computes the pairwise similarity within the same feature map so as to incorporate global information, whereas we compute the pairwise similarity between features of different images and use it to warp the corresponding color from the reference.

Given that the color warping is not accurate everywhere, we output the matching confidence map $\mathcal{S}$ indicating the reliability of sampling the reference color for each position $i$ of $x^l_t$:
		\begin{equation}
			\mathcal{S}(i) = \max_j \mathcal{M}(i,j)
		\end{equation}

In summary, our correspondence network generates two outputs: warped color $\mathcal{W}^{ab}$ and confidence map $\mathcal{S}$:
\begin{equation}
	(\mathcal{W}^{ab},\mathcal{S}) = \mathcal{N}(x^l_t,y^{lab}; \theta_{\mathcal{N}})
\end{equation}
       
		\paragraph{Colorization Subnet.}
		The correspondence is not accurate everywhere, so we employ the colorization network $\mathcal{C}$ which is parameterized by $\theta_\mathcal{C}$, to select the well-matched colors and propagate them properly. The network receives four inputs: the grayscale input $x^l_t$, the warped color map $\mathcal{W}^{ab}$ and the confidence map $\mathcal{S}$, and the colorized previous frame $\tilde{x}^{lab}_{t-1}$. Given these, this network outputs the predicted color map $\tilde{x}^{ab}_t$ for the current frame at $t$:
		\begin{equation}
			\tilde{x}^{ab}_t = \mathcal{C}(x^l_t,\mathcal{W}^{ab},\mathcal{S}|\tilde{x}^{lab}_{t-1}; \theta_{\mathcal{C}})
		\end{equation}
		Along with the luminance channel $x^l_t$, we obtain the colorized image $\tilde{x}^{lab}_t$, also denoted as $\tilde{x}_t$. 
		
		\subsection{Loss}
Our network is supposed to produce realistic video colorization without temporal flickering. Furthermore, the colorization style should resemble the reference in the corresponding regions. To accomplish these objectives, we impose the following losses. \vspace{-0.7em}

		\paragraph{Perceptual Loss.}
        First, to encourage the output to be perceptually plausible, we adopt the \textit{perceptual loss}~\cite{johnson2016perceptual} which measures the semantic difference between the output $\tilde{x}$ and the ground truth image $x$:
		\begin{equation}
			\mathcal{L}_{perc} = \Vert \Phi^L_{\tilde{x}} - \Phi^L_{x}\Vert ^2_2
		\end{equation}
		where $\Phi^L$ represent the feature maps extracted at the $relu{L\_2}$ layer from the VGG19 network. Here we set $L=5$ since the top layer captures mostly semantic information. This loss encourages the network to select the confident colors from $\mathcal{W}^{ab}$ and propagate them properly. \vspace{-0.7em}
        
        \paragraph{Contextual Loss.}
We introduce a \textit{contextual loss}, to encourage colors in the output to be close to those in the reference. The contextual loss is proposed in~\cite{mechrez2018contextual} to measure the local feature similarity while considering the context of the entire image, so it is suitable for transferring the color from the semantically related regions. Our work is the first to apply the contextual loss into exemplar-based colorization. The cosine distances $d^L(i,j)$ are first computed between each pair of feature points $\Phi^L_{\tilde{x}}(i)$ and $\Phi^L_y(j)$, and then normalized as $\tilde{d}^L (i,j) = d^L(i,j) / (\min_k d^L(i,k) + \epsilon), \epsilon=1e-5$. The pairwise affinities $A^L(i,j)$ between features are defined as: 

\begin{equation}
A^L(i,j) = \softmax_j(1-\tilde{d}^L(i,j)/h)
\end{equation}
		where we set the bandwidth parameter $h=0.1$ as a recommendation. The affinities $A^l(i,j)$ range within $[0,1]$ and measure the similarity of $\tilde{x}_t(i)$ and $y(j)$ with the $L$th layer features. Contrary to the backward matching in~\cite{mechrez2018contextual}, we use forward matching where for each feature $\Phi^l_{\tilde{x},i}$ we find the closest feature $\Phi^l_{y,j}$ in $y$. This is because some objects in $x^l_t$ may not exist in $y$. Consequently, the contextual loss is defined to maximize the affinities between the result and the reference:
		\begin{equation}
			\mathcal{L}_{context} = \sum_l w_{L}\left[-\log\left(\frac{1}{N_L}\sum_i \max_j A^L(i,j)\right) \right].
		\end{equation}
		Here we use multiple feature maps: $L=2$ to $5$. $N_L$ denotes the feature number of layer $L$.  We set higher weights $w_{L}$ for higher level features as the correspondence is proven more reliable using the coarse-to-fine searching strategy~\cite{liao2017visual}. \vspace{-0.7em}
		
	   \paragraph{Smoothness Loss.}
		We introduce a \textit{smoothness loss} to encourage spatial smoothness. We assume that neighboring pixels of $\tilde{x}_t$ should be similar if they have similar chrominance in the ground truth image $x_t$. The smoothness loss is defined as the difference between the color of current pixel and the weighted color of its 8-connected neighborhoods:
		\begin{equation}
			\begin{split}
			\mathcal{L}_{smooth} = &\frac{1}{N} \sum_{c\in\{a,b\}} \sum_i \left(\tilde{x}^{c}_t(i) - \sum_{j\in \mathbb{N}(i)} w_{i,j} \tilde{x}_t^{c}(j)\right) 
			\end{split}
		\end{equation}
		where $w_{i,j}$ is the WLS weight~\cite{farbman2008edge} which measures the neighborhood correlations. This edge-aware weight helps to produce edge-preserving colorization and alleviate color bleeding artifacts. \vspace{-0.7em}

        \paragraph{Adversarial Loss.}
		We also employ an \textit{adversarial loss} to constrain the colorization video frames to remain
		realistic. Instead of using image discriminator, a video discriminator is used to evaluate consecutive video frames. We assume that flickering and defective videos can be easily distinguished from real ones, so the colorization network can learn to generate coherent natural results during the adversarial training.
		
		It is difficult to stabilize the adversarial training especially on a large-scale dataset like ImageNet. In this work we adopt the relativistic discriminator~\cite{jolicoeur2018relativistic} which estimates the extent in which the real frames (denoted as $z_{t-1}$ and $z_{t}$) look more realistic than the colorized ones $\tilde{x}_{t-1}$ and $\tilde{x}_{t}$. We adopt the least squares GAN in its relativistic format and the loss for the generator $G$ is defined as:\vspace{-0.7em}
		
		\begin{equation}
			\begin{split}
			\mathcal{L}_{adv}^G = \ &\mathbb{E}_{(\tilde{x}_{t-1},\tilde{x}_{t})\sim \mathcal{P}_{\tilde{x}}}[(D(\tilde{x}_{t-1},\tilde{x}_{t}) \\&- \mathbb{E}_{(z_{t-1} z_t)\sim \mathcal{P}_{z}}D(z_{t-1},z_t) - 1)^2] \\
			& + \mathbb{E}_{(z_{t-1} z_t)\sim \mathcal{P}_z}[ (D(z_{t-1}, z_t) \\
			& - \mathbb{E}_{(\tilde{x}_{t-1},\tilde{x}_{t})\sim \mathcal{P}_{\tilde{x}}}D(\tilde{x}_{t-1},\tilde{x}_{t}) +1)^2]
			\end{split}
		\end{equation}
		 The relative discriminator loss can be defined in a similar way. From our experiments, this GAN is better to stabilize training than a standard GAN. \vspace{-1.0em}
		

    \paragraph{Temporal Consistency Loss.}
    To efficiently consider temporal coherency, we also impose a \textit{temporal consistency loss}~\cite{chen2017coherent} which explicitly penalizes the color change along the flow trajectory:
	\begin{equation}
	\begin{split}
	    \mathcal{L}_{temporal} = \Vert m_{t-1}\odot W_{t-1, t}(\tilde{x}_{t-1}^{ab}) - m_{t-1}\odot\tilde{x}_t^{ab}\Vert
	\end{split}
	\label{eq:flow}
	\end{equation}
	where $W_{t-1, t}$ is the forward flow from the last frame $x_{t-1}$ to $x_{t}$ and $m_{t-1}$ is the binary mask which excludes the occlusion, and $\odot$ represents the Hadamard product.

    \paragraph{L1 Loss.}
    With the above loss functions, the network can already generate high quality plausible colorized results given a customized reference. Still, we want the network degenerate to the case where the reference comes from the same scene as the video frames. This is a common case for video colorization applications. In this case, we have the ground truth of the predicted frame, so add one more \textit{L1 loss} term to measure the color difference between output $\tilde{x}_{t}$ and the ground truth $x_t$:   
	 \begin{equation}
		\mathcal{L}_{L1} = \Vert \tilde{x}_{t}^{ab} - x_t^{ab}\Vert_1
		\label{eq:l1}
	\end{equation}\vspace{-1.5em}

	\paragraph{Objective Function.}
	Combining all the above losses, and the overall objective we aim to optimize is:
		\begin{equation}
			\begin{split}
			\mathcal{L}_{I} = & \lambda_{perc}\mathcal{L}_{perc} + \lambda_{context}\mathcal{L}_{context} +
			\lambda_{smooth}\mathcal{L}_{smooth}\\&+ \lambda_{adv}\mathcal{L}_{adv} 
			+ \lambda_{temporal}\mathcal{L}_{temporal} 
		    + \lambda_{L1}\mathcal{L}_{L1}
			\end{split}
		\end{equation}
		where $\lambda$ controls the relative importance of terms. With the guidance of these losses, we successfully unify the correspondence and color propagation within a single network, which learns to generate plausible results based on the exemplar image.

	\section{Implementation}
    \paragraph{Network Structure.}
	The correspondence network involves 4 residual blocks each with 2 \textit{conv} layers. The colorization subnet adopts an auto-encoder structure with skip-connections to reuse the low-level features. There are 3 convolutional blocks in the contractive encoder and 3 convolutional blocks in the decoder which recovers the resolution; each convolutional block contains 2$\sim$3 \textit{conv} layers. The $tanh$ serves as the last layer to bound the chrominance output within the color space. The video discriminator consists of 7 \textit{conv} layers where the first six layers halve the input resolution progressively. Also, we insert the self-attention block~\cite{zhang2018self} after the second \textit{conv} layer to let the discriminator examine the global consistency. We use instance normalization since colorization should not be affected by the samples in the same batch. To further improve training stability we apply spectral normalization~\cite{miyato2018spectral} on both generator and discriminator as suggested in~\cite{zhang2018self}.

   \paragraph{Training.}
   \begin{figure}[!tb]
	\setlength\tabcolsep{0pt}
	\begin{center}
    \begin{tabular}{ccc}
    \includegraphics[width=0.333\columnwidth]{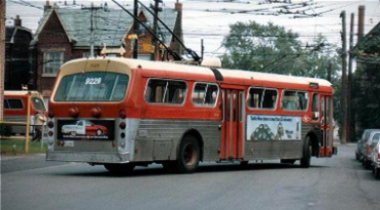}&
    \includegraphics[width=0.333\columnwidth]{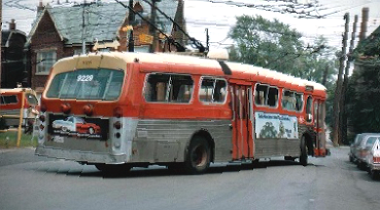}&
    \includegraphics[width=0.333\columnwidth]{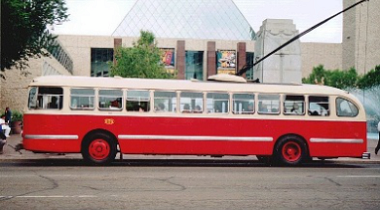}\\
    \end{tabular}
    \end{center}
    \caption{Augmented training images from ImageNet dataset.}
    \label{fig:training_image}
  	\end{figure}
  	
	In order to cover a wide range of scenes, we use multiple datasets for training. First, we collect 1052 videos from Videvo stock~\cite{Videvo} which mainly contains animals and landscapes. Furthermore, we include more portraits videos using the Hollywood2 dataset~\cite{marszalek09}. We filter out the videos that are either too dark or too faded in color, leaving 768 videos for training. For each video clip we provide reference candidates by inquiring the five most similar images from the corresponding class in the ImageNet dataset. We extract 25 frames from each video and use FlowNet2~\cite{ilg2017flownet} to compute the optical flow required for the temporal consistency loss and use the method~\cite{ruder2016artistic} for the occlusion mask. To further expand the data category, we include images in the ImageNet and apply random geometric distortion and luminance noises to generate augmented video frames as shown in Figure~\ref{fig:training_image}. Thus, we get 70k augmented videos in diverse categories. To suit the standard aspect ratio 16:9, we crop all the training images to $384\times 216$. We occasionally provide the reference which is the ground truth image itself but insert Gaussian noise, or feature noise to the VGG19 features before feeding them into the correspondence network. We deliberately cripple the color matching during training, so the colorization network better learns the color propagation even when the correspondence is inaccurate.  
    
    We set $\lambda_{perc} = 0.001$, $\lambda_{context}=0.2$, $\lambda_{smooth}=5.0$, $\lambda_{adv}=0.2$, $\lambda_{flow}=0.02$ and $\lambda_{L1} = 2.0$. We use a learning rate of $2\times 10^{-4}$ for both generator and discriminator without any decay schedule and train the network using the AMSGrad solver with parameters $\beta_1 = 0.5$ and $\beta_2=0.999$. We train the network for 10 epochs with a batch size of 40 pairs of video frames.

	\section{Experiments}
	In this section, we first study the effectiveness of individual components in our method. Then, we compare our method with state-of-the-art approaches. 
	\subsection{Ablation Studies}
\begin{figure}[!tb]
    \setlength\tabcolsep{1.5pt}
    \centering
    \small
    \begin{tabular}{ccc}
    \raisebox{0\height}{Input images} & \raisebox{0\height}{Warped color image} & \raisebox{0\height}{Colorized result}\\
    \includegraphics[width=0.3\columnwidth]{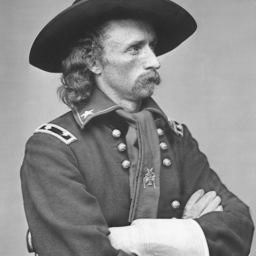}&
    \includegraphics[width=0.3\columnwidth]{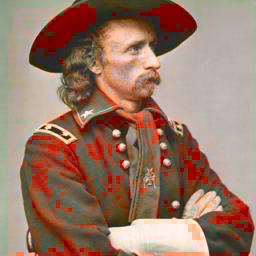}&
    \includegraphics[width=0.3\columnwidth]{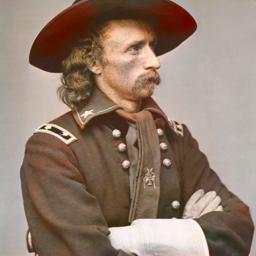}\\
    \includegraphics[width=0.3\columnwidth]{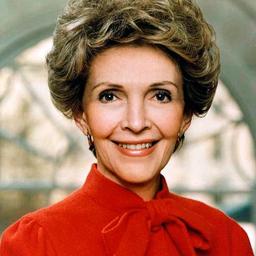}&
    \includegraphics[width=0.3\columnwidth]{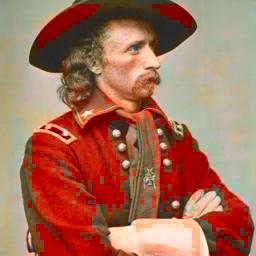}&
    \includegraphics[width=0.3\columnwidth]{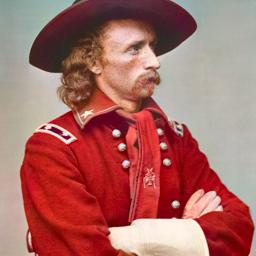}
    \end{tabular}
    \caption{First row: nearest neighbor matching. Second row: with learning parameters in the correspondence network. The first columns are the grayscale image and reference image respectively.}
    \label{fig:learn_warping}
    \vspace{-0.5em}
  	\end{figure}

	\paragraph{Correspondence Learning.}
	To demonstrate the importance of learning parameters in the correspondence subnet, we compare our method with nearest neighbor (NN) matching, in which each feature point of the input image will be matched to the nearest neighbor of the reference feature. Figure~\ref{fig:learn_warping} shows that our learning-based method matches mostly correct colors from the reference and eases color propagation for the colorization subnet.\vspace{-0.7em}
	
	\begin{figure*}[!htb]
    \setlength\tabcolsep{1pt}
    \centering
    \small
    \begin{tabular}{ccccccc}
    Input image & Reference & w/o $\mathcal{L}_{perc}$ & w/o $\mathcal{L}_{context}$ & w/o $\mathcal{L}_{smooth}$ & w/o $\mathcal{L}_{adv}$ & Full  \\
    \includegraphics[width=0.285\columnwidth]{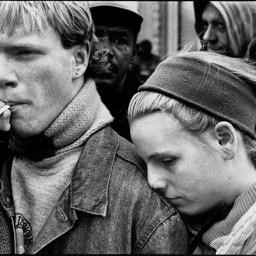}&
    \includegraphics[width=0.285\columnwidth]{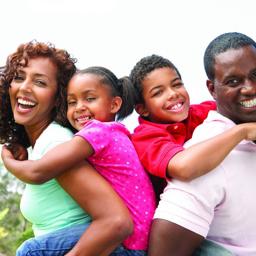}&
    \includegraphics[width=0.285\columnwidth]{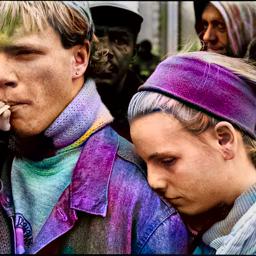}&
    \includegraphics[width=0.285\columnwidth]{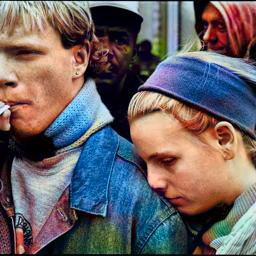}&
    \includegraphics[width=0.285\columnwidth]{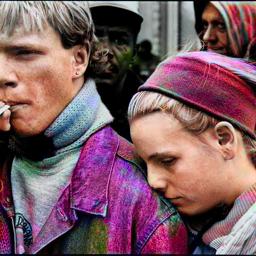}&
    \includegraphics[width=0.285\columnwidth]{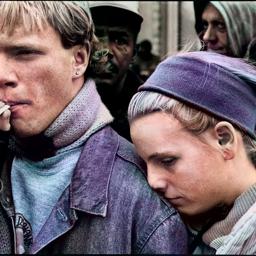}&
    \includegraphics[width=0.285\columnwidth]{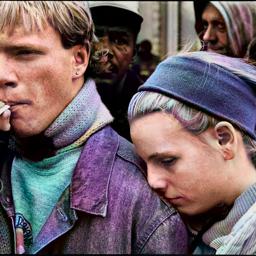}
    \end{tabular}
    \caption{Ablation study for different loss functions. Please refer to the supplementary material for the quantitative comparisons.}
    \label{fig:ablation}
  	\end{figure*}

	
	\begin{figure*}[!htb]
    \setlength\tabcolsep{1.5pt}
    \small
    \begin{center}
    \begin{tabularx}{\textwidth}{ccccccc}
Input & Reference & \textbf{Ours} & \cite{he2018deep} & \cite{iizuka2016let} & \cite{larsson2016learning} & \cite{zhang2016colorful} \\
    \includegraphics[width=0.135\textwidth]{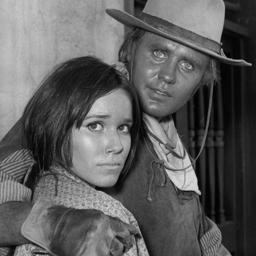}&
    \includegraphics[width=0.135\textwidth]{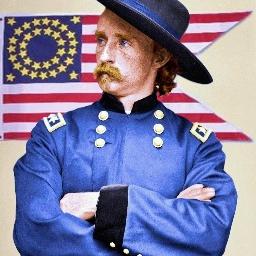}&
    \includegraphics[width=0.135\textwidth]{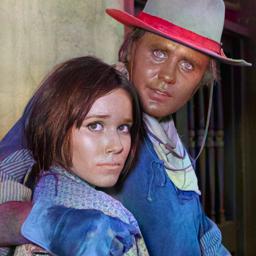}&
    \includegraphics[width=0.135\textwidth]{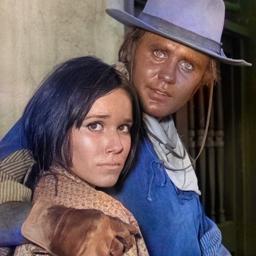}&
    \includegraphics[width=0.135\textwidth]{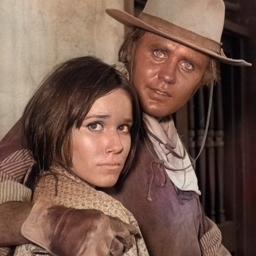}&
    \includegraphics[width=0.135\textwidth]{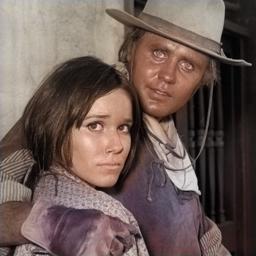}&
    \includegraphics[width=0.135\textwidth]{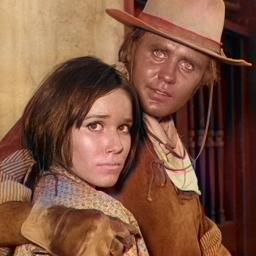}\\
    
    \includegraphics[width=0.135\textwidth]{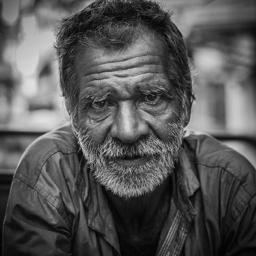}&
    \includegraphics[width=0.135\textwidth]{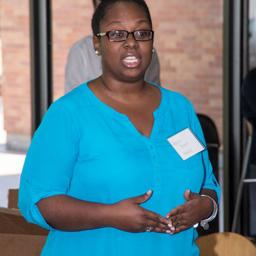}&
    \includegraphics[width=0.135\textwidth]{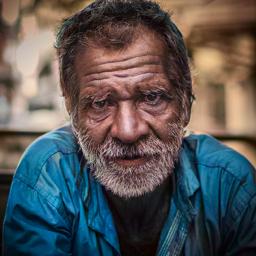}&
    \includegraphics[width=0.135\textwidth]{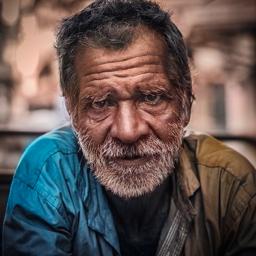}&
    \includegraphics[width=0.135\textwidth]{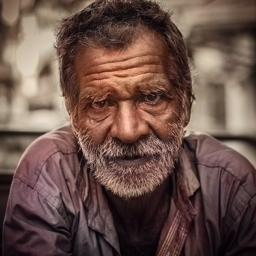}&
    \includegraphics[width=0.135\textwidth]{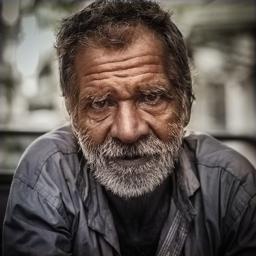}&
    \includegraphics[width=0.135\textwidth]{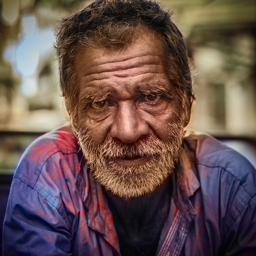}\\

    \includegraphics[width=0.135\textwidth]{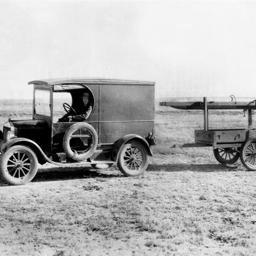}&
    \includegraphics[width=0.135\textwidth]{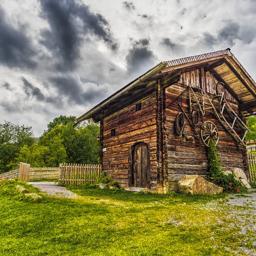}&
    \includegraphics[width=0.135\textwidth]{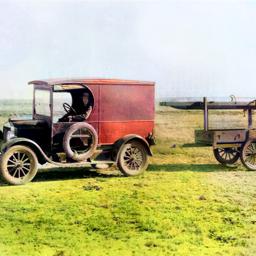}&
    \includegraphics[width=0.135\textwidth]{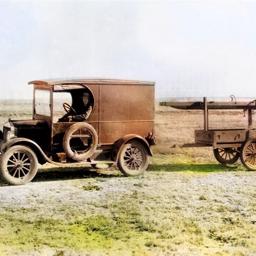}&
    \includegraphics[width=0.135\textwidth]{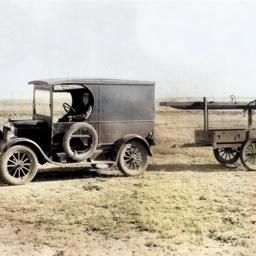}&
    \includegraphics[width=0.135\textwidth]{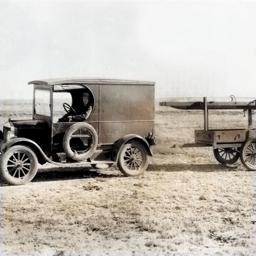}&
    \includegraphics[width=0.135\textwidth]{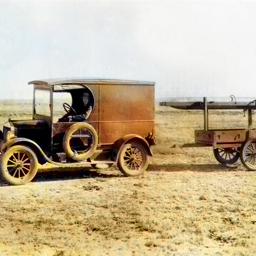}
    \end{tabularx}
    \end{center}
    \vspace{-0.5em}
    \caption{Comparison with image colorization with state-of-the-art methods.}
    \label{fig:image_comparison}
    \vspace{-0.5em}
  	\end{figure*}

  	\begin{figure}[!tb]
	\centering
	\includegraphics[width=0.95\linewidth]{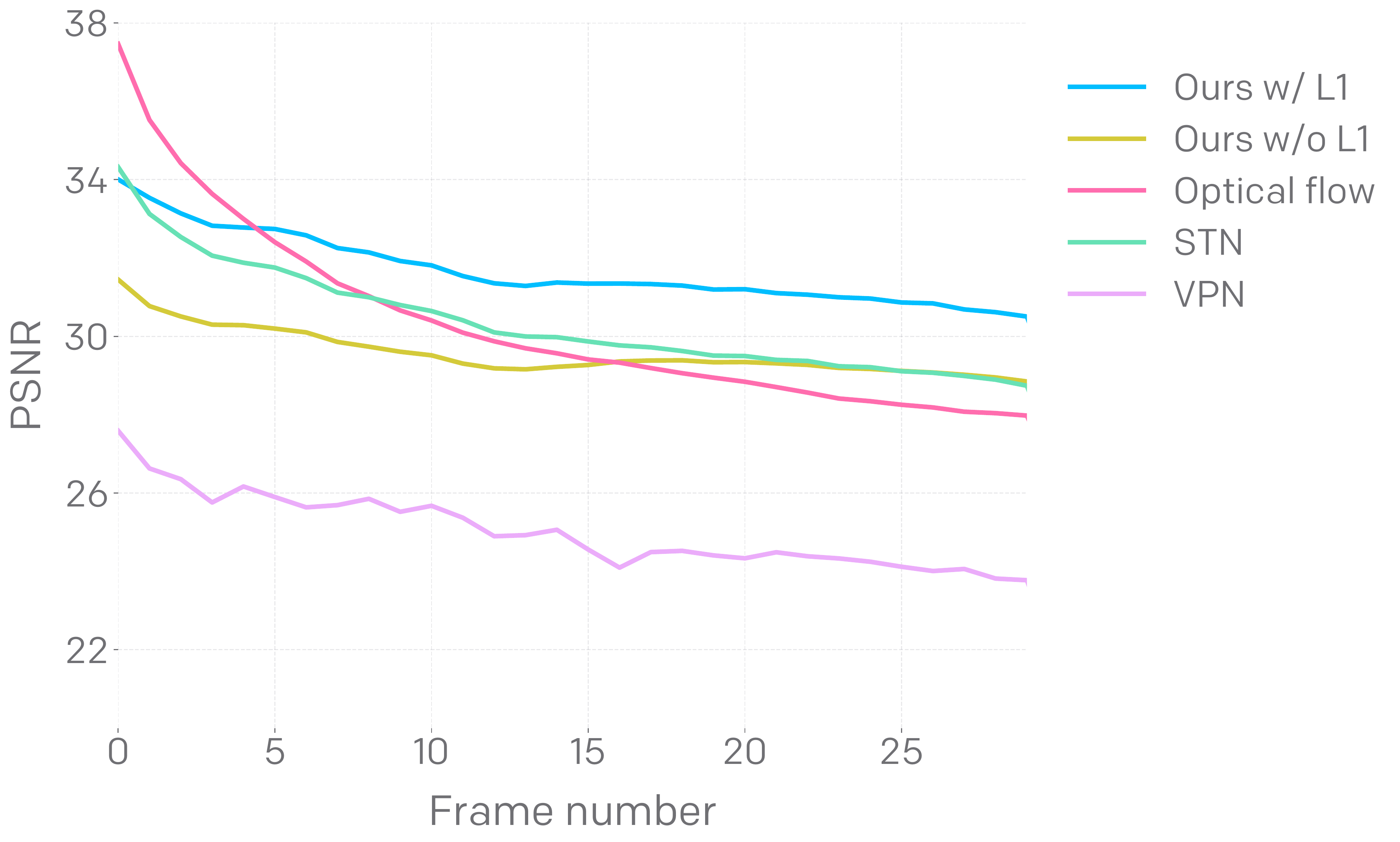}
	\caption{Quantitative comparison with video color propagation.}
	\label{fig:propagation_curve}
	\end{figure}
	
	 \begin{figure}[t]
	\centering
	\includegraphics[width=0.99\linewidth]{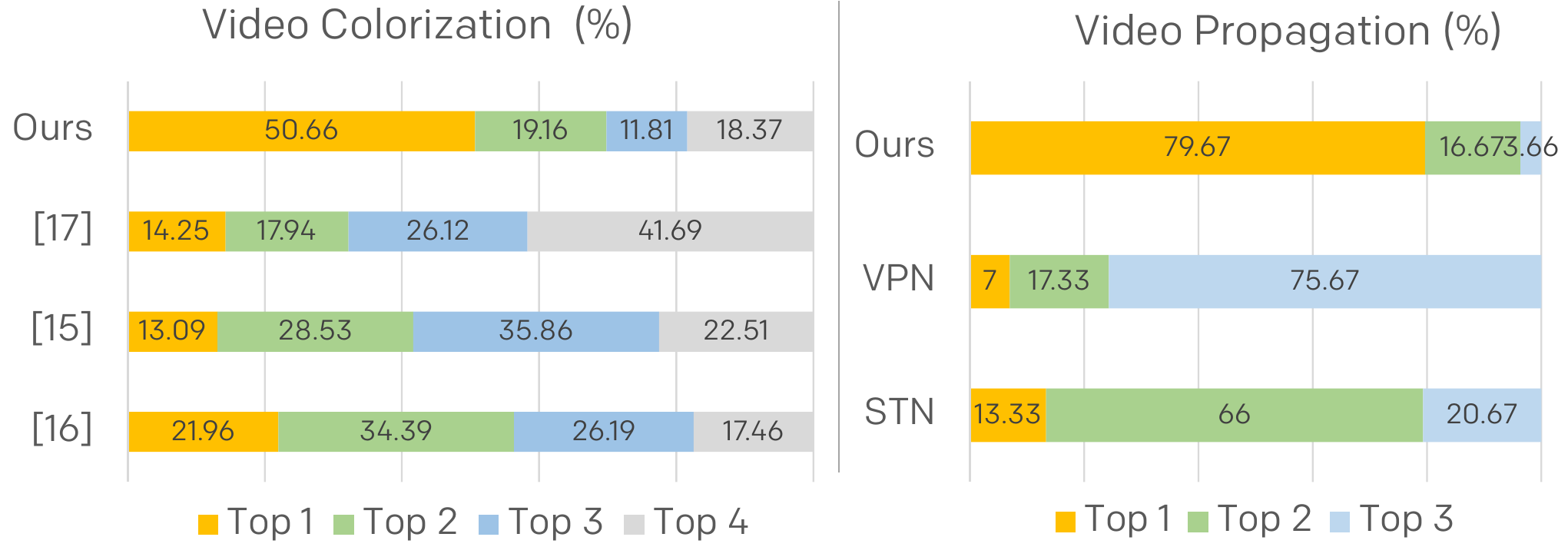}
	\caption{User study results.}
	\label{fig:user_study}
	\end{figure}
  	
  	\paragraph{Analysis of Loss Functions.}
	We ablate the loss functions individually and evaluate their importance, as shown in Figure~\ref{fig:ablation}. When we remove $\mathcal{L}_{perc}$, the colorization fully adopts the color from the reference, but tends to produce more artifacts since there is no loss function to constrain the output to be semantically similar to the input. When we remove $\mathcal{L}_{context}$, the output does not resemble the reference style. When $\mathcal{L}_{smooth}$ is ablated, colors may not be fully propagated to the whole coherent region. Without $\mathcal{L}_{adv}$, the color appears washed out and unrealistic. This is because color warping is not accurate and the final output becomes the local color average of the warping colors. In comparison, our full model produces vivid colorization with fewer artifacts.
	
  	
	\subsection{Comparisons}
	\paragraph{Comparison with Image Colorization.}

  	\begin{table}[t]
	\small
	\centering
    {\setlength\tabcolsep{0pt}}%
	\begin{tabularx}{\linewidth}{LLLLLLL} 
	\hline
	 & Top-5 Acc(\%) & Top-1 Acc(\%) & FID  & Colorful  & Flicker \\
	\hline
	GT & 90.27 & 71.19 & 0.00 & 19.1& 5.22 \\
	~\cite{iizuka2016let} & 85.03 & 62.94 &7.04&11.17& 7.19/5.69+\\
	~\cite{larsson2016learning} & 84.76 & 62.53 &7.26&10.47&6.76/5.42+\\
	~\cite{zhang2016colorful} & 83.88 & 60.34 &8.38&\textbf{20.16}&7.93/5.89+\\
	~\cite{he2018deep} & 85.08 & 64.05 & 4.78& 15.63 & NA\\
	Ours &\textbf{85.82} &\textbf{64.64}  &\textbf{4.02}&17.90 & 5.84\\
	\hline
	\end{tabularx}
	\vspace{0.1em}
	\caption{Comparison with image and per-frame video colorization methods (image test dataset: ImageNet 10k and video test dataset: Videvo.)}
	\label{table:comparison_image}
	\vspace{-.8em}
	\end{table}
	
	
	We compare our method against recent learning based image colorization methods both quantitatively and qualitatively. The baseline methods include three automatic colorization methods (Iizuka et al.~\cite{iizuka2016let}, Larsson et al.~\cite{larsson2016learning} and Zhang et al.~\cite{zhang2016colorful}) and one exemplar based method (He and Chen et al.~\cite{he2018deep}) since these methods are regarded as state-of-the-art. 
	
	For the quantitative comparison, we test these methods on 10k subset of the ImageNet dataset, as shown in Table~\ref{table:comparison_image}. For exemplar based methods, we take the Top-1 recommendation from ImageNet as the reference. First, we measure the classification accuracy using the VGG19 pre-trained on color images. Our method gives the best \textit{Top-5} and \textit{Top-1} class accuracy, indicating that our method produces semantically meaningful results. Second, we employ the \textit{Fr\'echet Inception Distance} (FID)~\cite{heusel2017gans} to measure the semantic distance between the colorized output and the realistic natural images. Our method achieves the lowest FID, showing that our method provides the most realistic results. In addition, we measure the colorfulness using the psychophysics metric from~\cite{hasler2003measuring} due to the fact that the users usually prefer colorful images. Table~\ref{table:comparison_image} shows that Zhang et al.'s work~\cite{zhang2016colorful} produces the most vivid color since it encourages rare colors in the loss function; however their method tends to produce visual artifacts, which are also reflected in FID score and the user study. Overall, the results of our method, though slightly less vibrant, exhibit similar colorfulness to the ground truth. The qualitative comparison (in Figure~\ref{fig:image_comparison}) also indicates that our method produces the most realistic, vibrant colorization results.\vspace{-0.7em}
	
  	\begin{figure*}[!tb]
    \setlength\tabcolsep{1.5pt}
    \centering
    \small
    \begin{tabularx}{\textwidth}{ccccc}
    &$T=0$ & $T=15$ & $ T=30$ & $T=45$ \\
    \raisebox{0.145\height}{\rotatebox{90}{Ground truth}}&
    \includegraphics[width=0.22\textwidth]{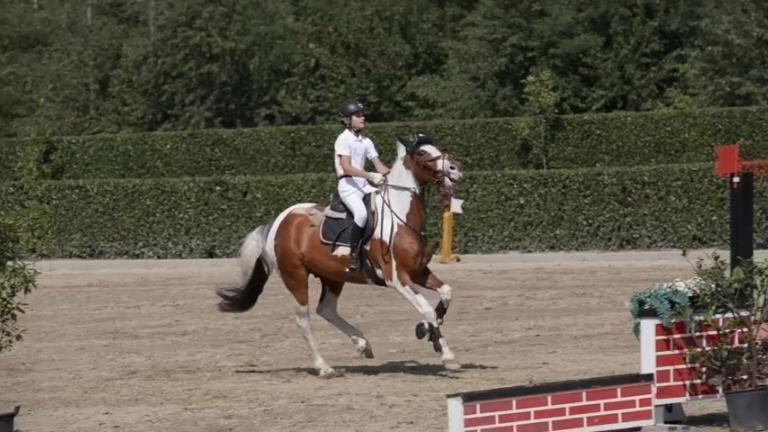}&
    \includegraphics[width=0.22\textwidth]{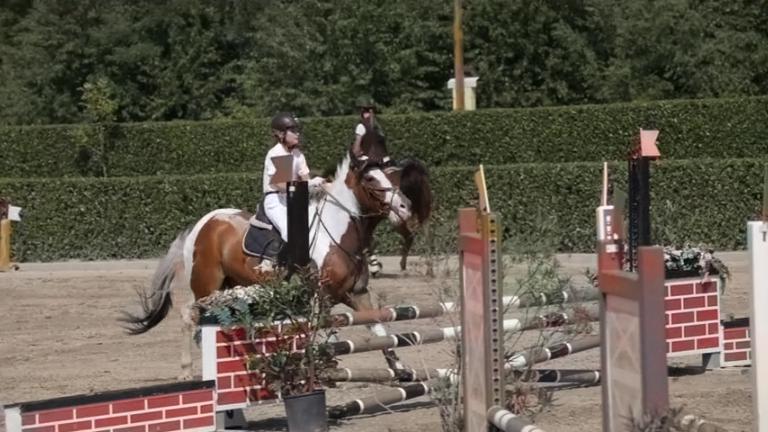}&
    \includegraphics[width=0.22\textwidth]{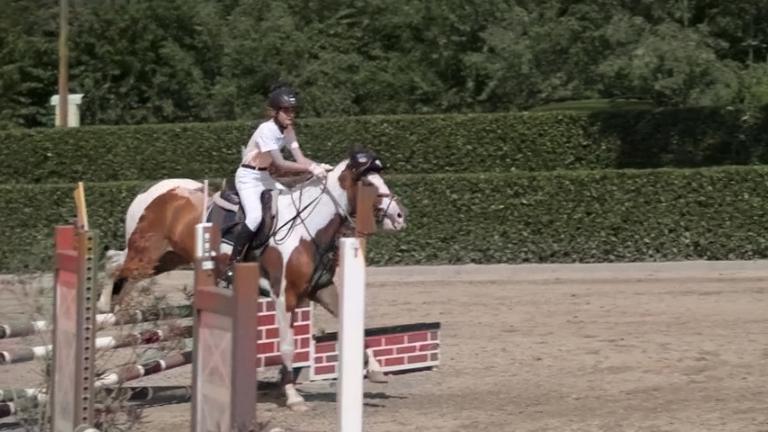}&
    \includegraphics[width=0.22\textwidth]{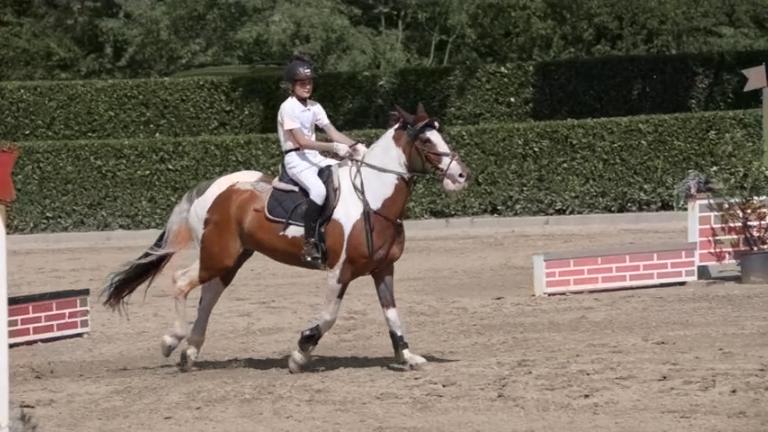}\\
    \raisebox{1.25\height}{\rotatebox{90}{VPN}}&
    \includegraphics[width=0.22\textwidth]{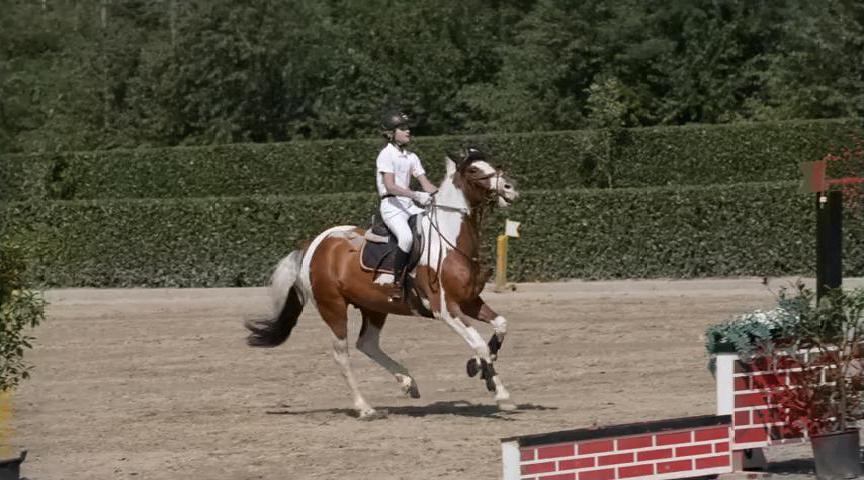}&
    \includegraphics[width=0.22\textwidth]{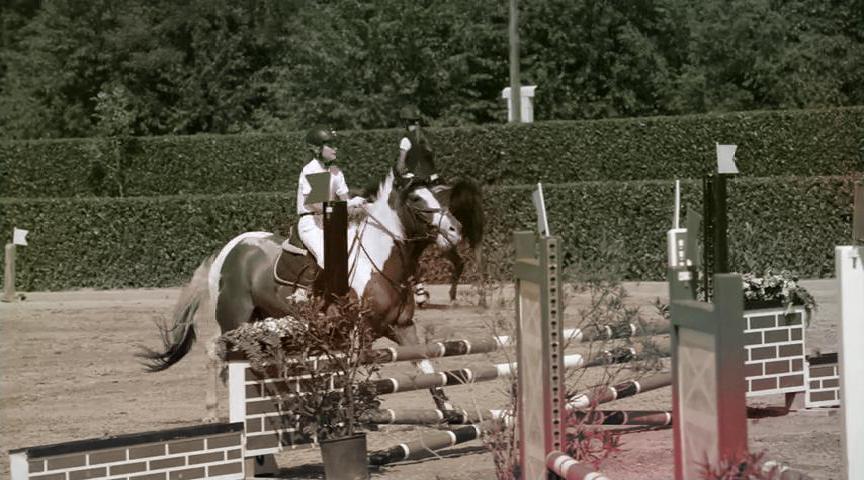}&
    \includegraphics[width=0.22\textwidth]{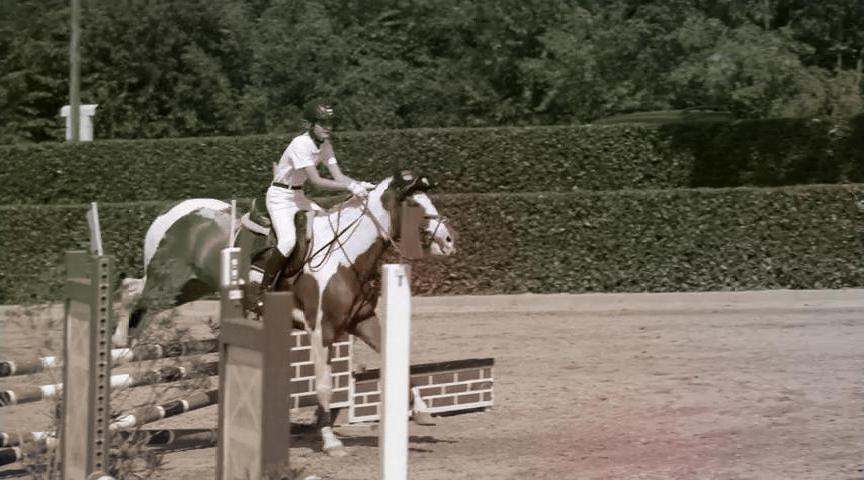}&
    \includegraphics[width=0.22\textwidth]{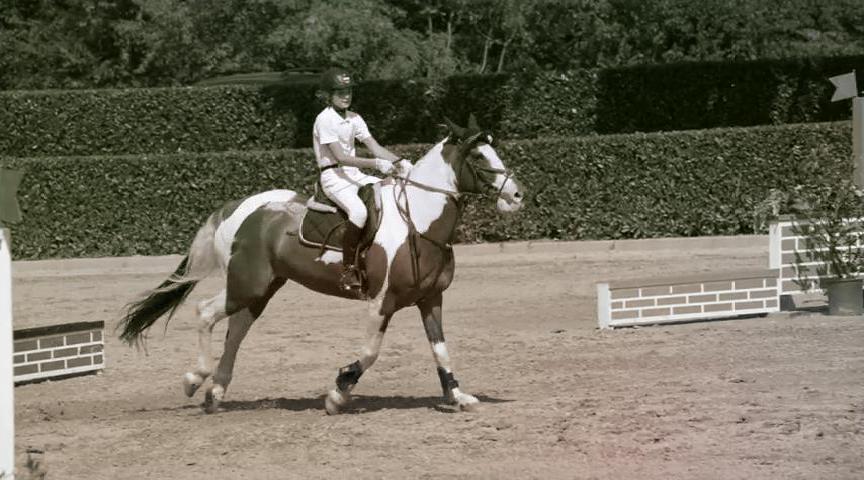}\\
    \raisebox{1.28\height}{\rotatebox{90}{STN}}&
    \includegraphics[width=0.22\textwidth]{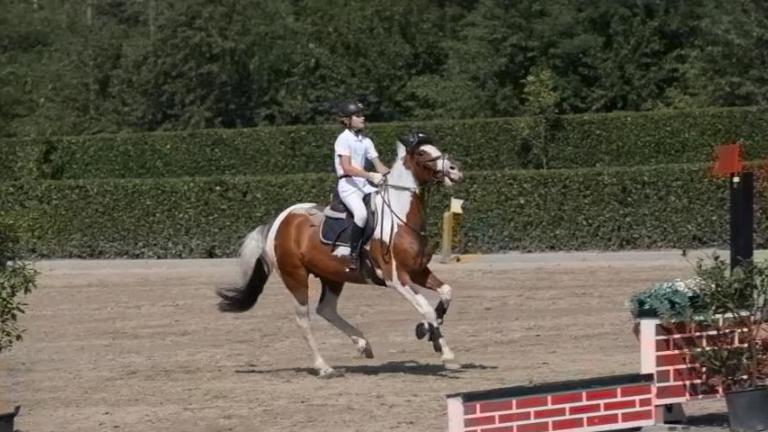}&
    \includegraphics[width=0.22\textwidth]{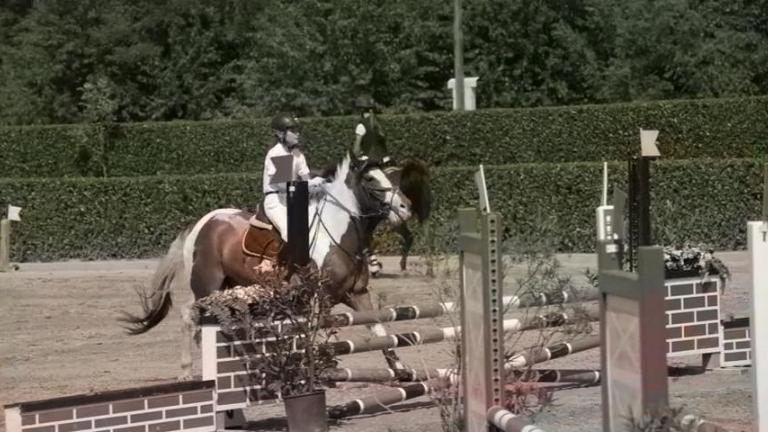}&
    \includegraphics[width=0.22\textwidth]{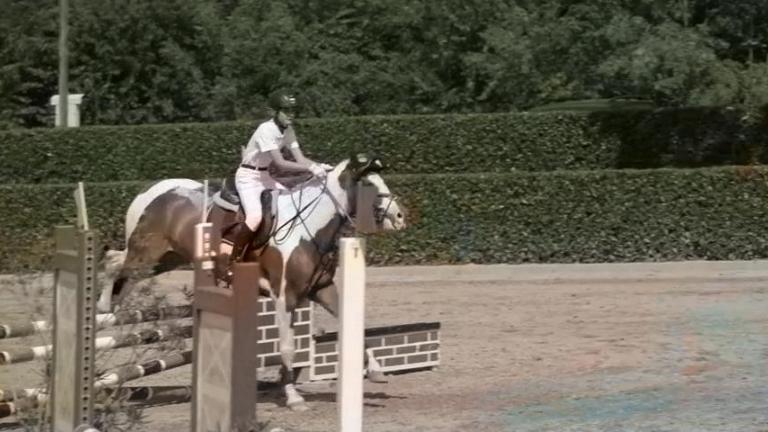}&
    \includegraphics[width=0.22\textwidth]{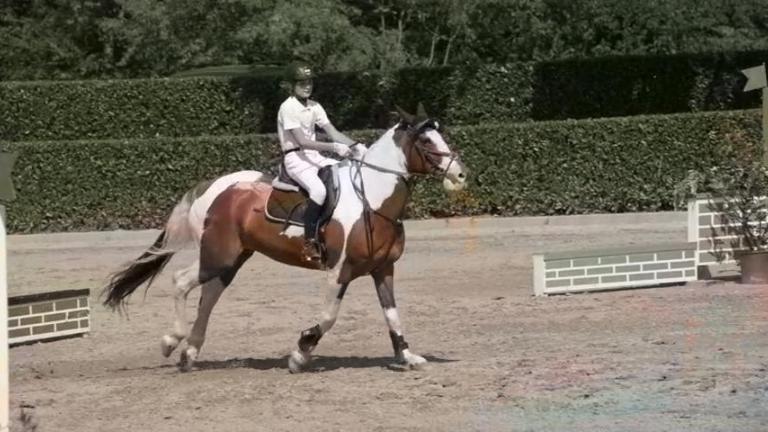}\\
    \raisebox{1.0\height}{\rotatebox{90}{\textbf{Ours}}}&
    \includegraphics[width=0.22\textwidth]{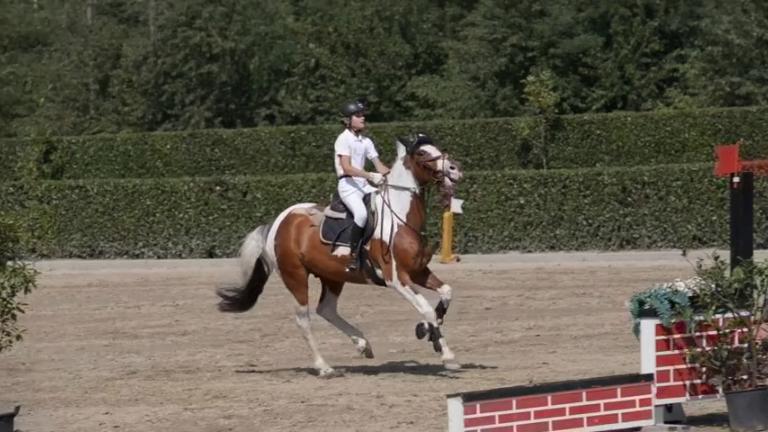}&
    \includegraphics[width=0.22\textwidth]{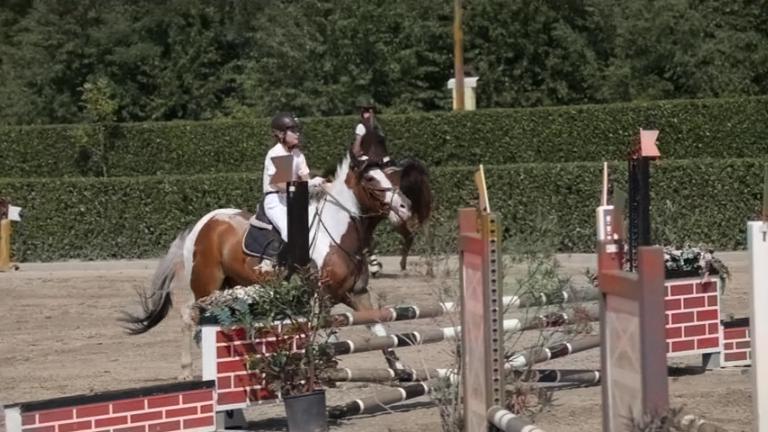}&
    \includegraphics[width=0.22\textwidth]{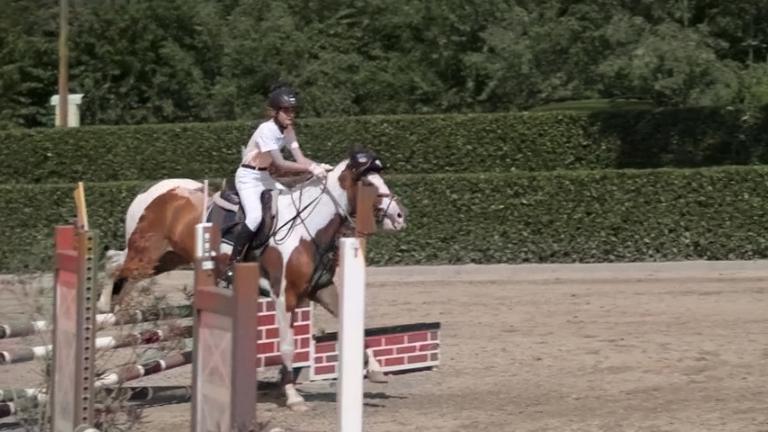}&
    \includegraphics[width=0.22\textwidth]{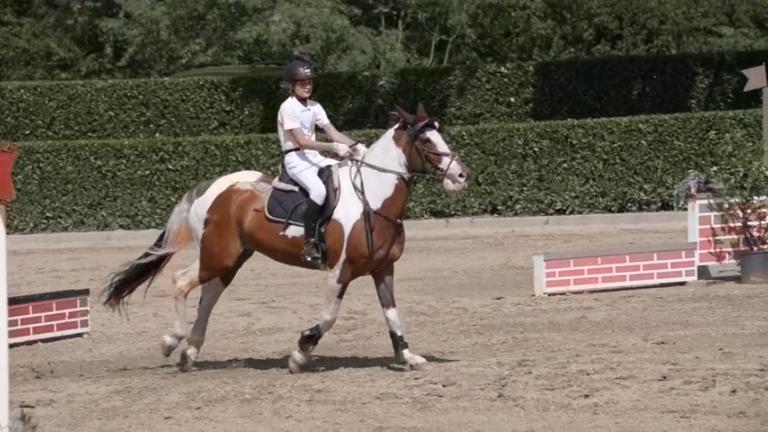}
    \end{tabularx}
    \vspace{0.1em}
    \caption{Comparison with video color propagation. With a given color frame as start, colors are propagated to the succeeding video frames. While other methods purely rely on color propagation, our method takes the initial color frame as a reference and is able to propagate colors for longer interval.}
    \label{fig:video_comparison_propagation1}
    \vspace{-0.5em}
  	\end{figure*}
  	
  	\begin{figure}[!tb]
    \setlength\tabcolsep{1.5pt}
    \begin{center}
    \small
    \begin{tabularx}{\textwidth}{cccc}
    \raisebox{0.8\height}{\rotatebox{90}{{\cite{larsson2016learning}}}}&
    \includegraphics[width=0.145\textwidth]{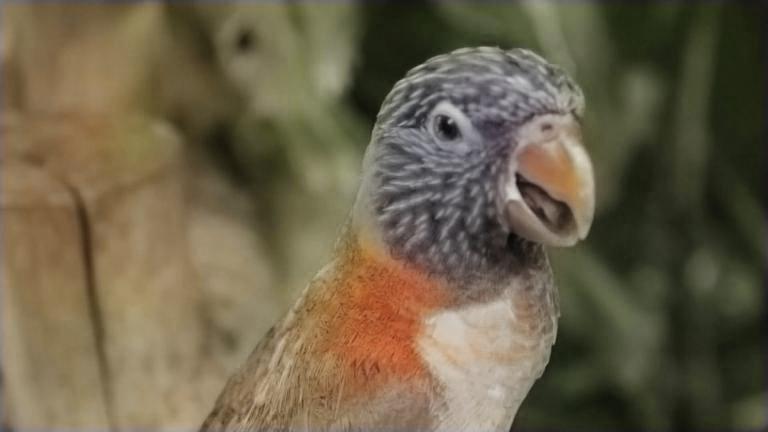}&
    \includegraphics[width=0.145\textwidth]{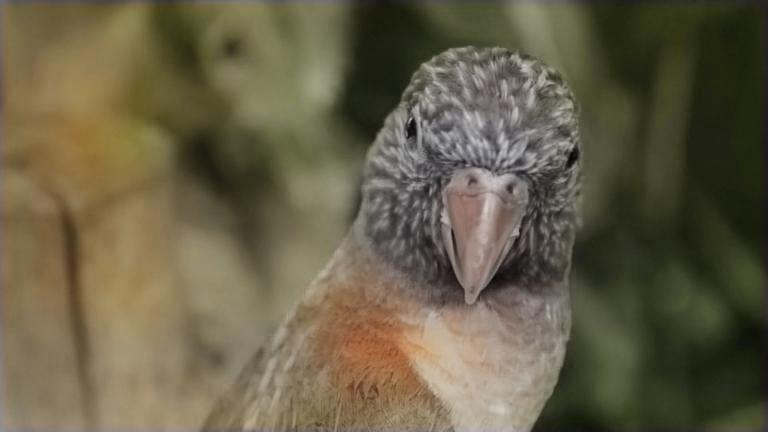}&
    \includegraphics[width=0.145\textwidth]{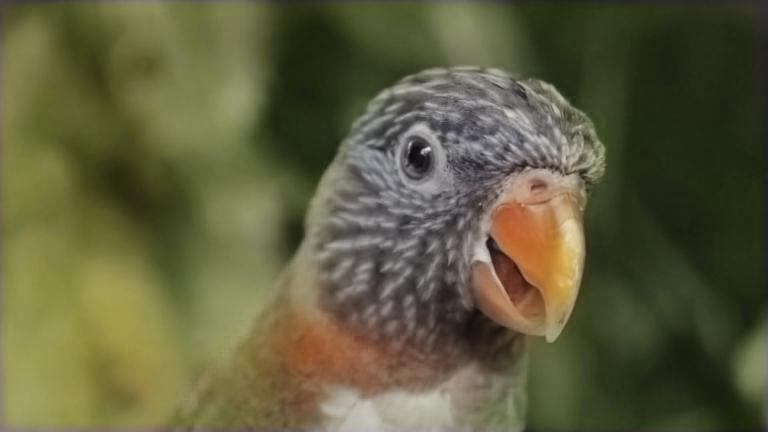}
    \\
    \raisebox{0.8\height}{\rotatebox{90}{{\cite{iizuka2016let}}}}&
    \includegraphics[width=0.145\textwidth]{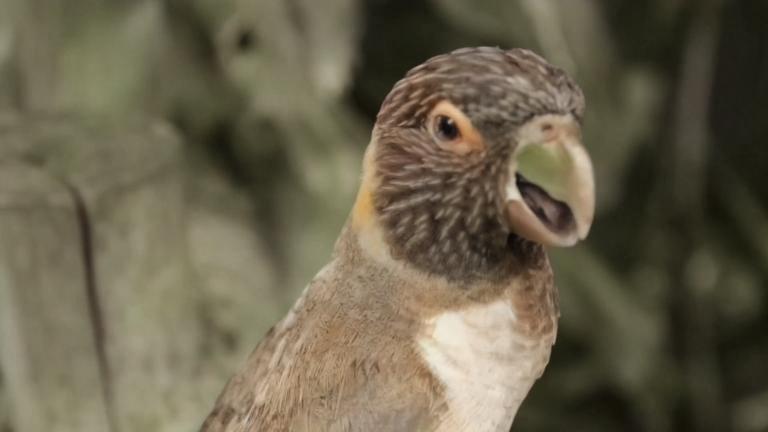}&
    \includegraphics[width=0.145\textwidth]{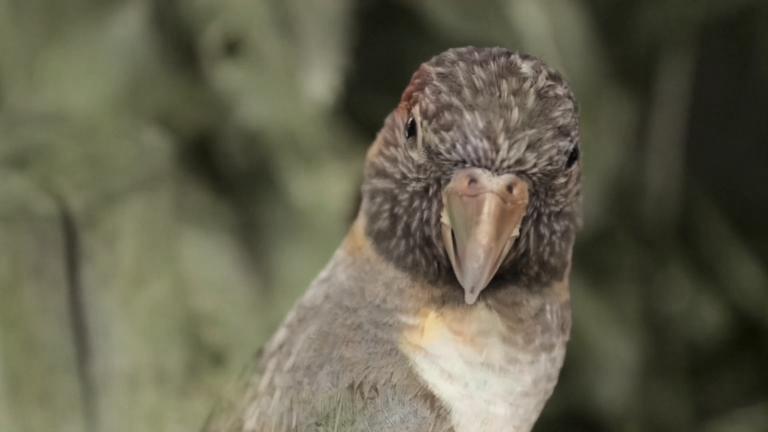}&
    \includegraphics[width=0.145\textwidth]{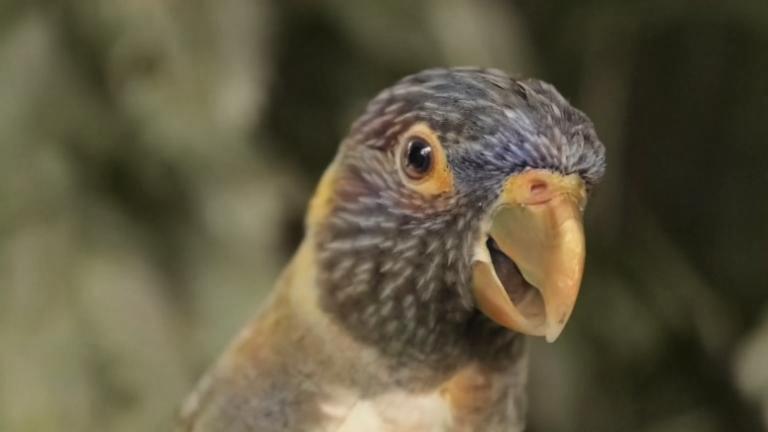}
    \\
    \raisebox{0.8\height}{\rotatebox{90}{{\cite{zhang2016colorful}}}}&    \includegraphics[width=0.145\textwidth]{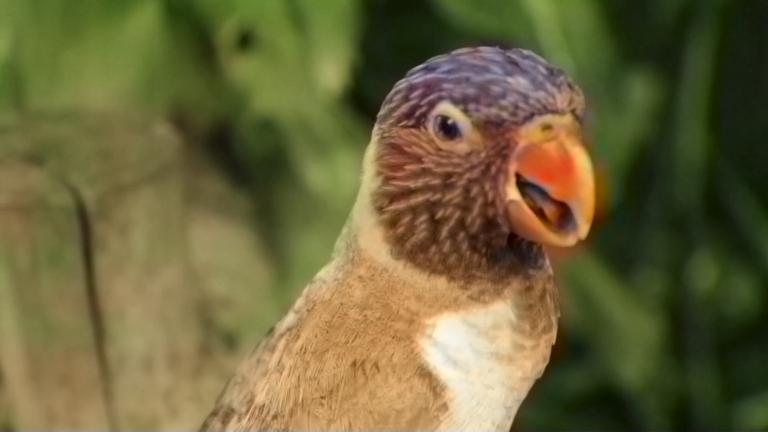}&
    \includegraphics[width=0.145\textwidth]{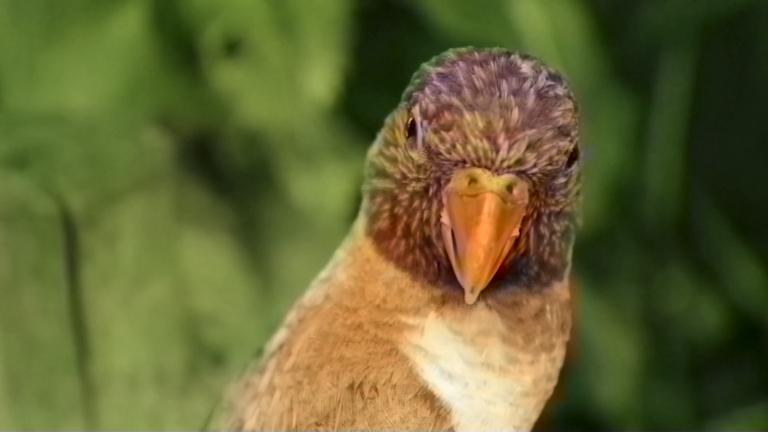}&
    \includegraphics[width=0.145\textwidth]{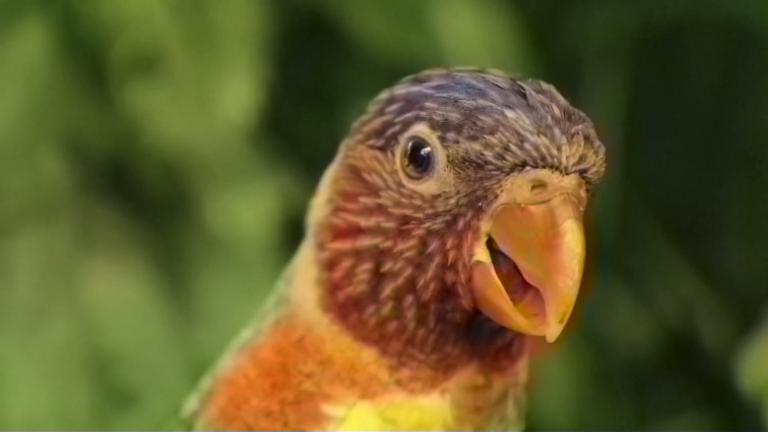}
    \\
    \raisebox{0.55\height}{\rotatebox{90}{{\textbf{Ours}}}}&
    \includegraphics[width=0.145\textwidth]{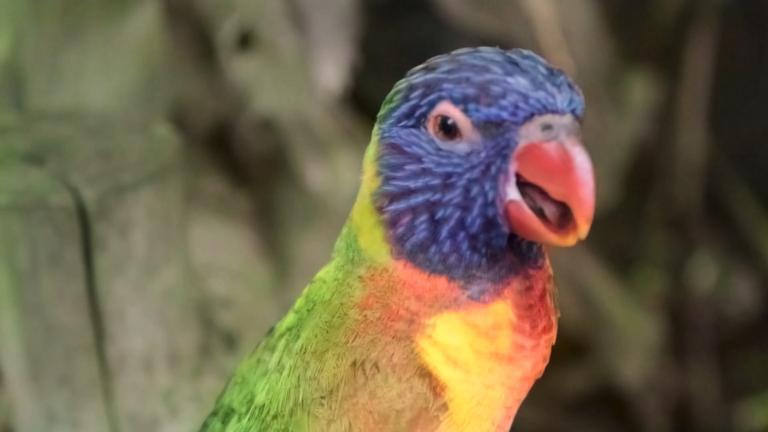}&
    \includegraphics[width=0.145\textwidth]{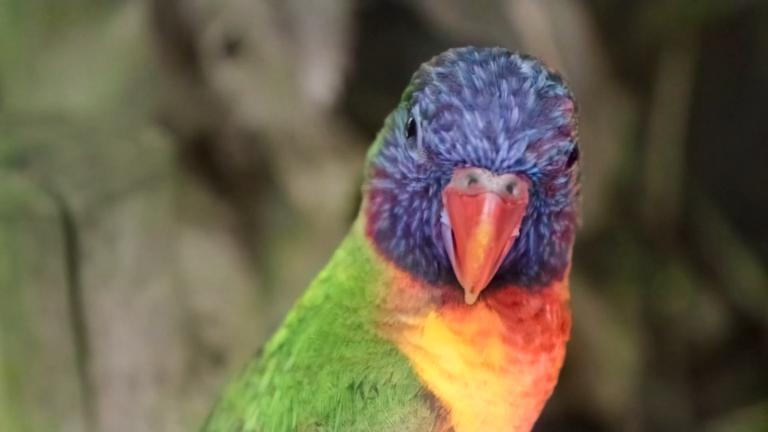}&
    \includegraphics[width=0.145\textwidth]{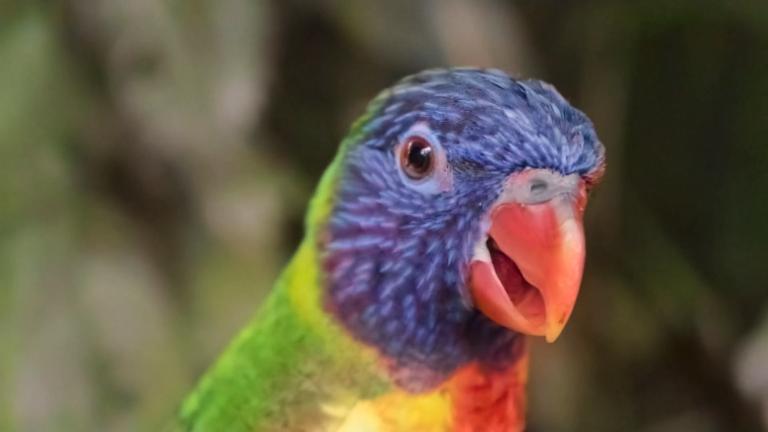}
    \\
    \end{tabularx}
    \end{center}
    \caption{Comparison with automatic video colorization.}
    \label{fig:video_comparison1}
    \vspace{-0.8em}
  	\end{figure}

	\paragraph{Comparison with Automatic Video Colorization.}
	In this experiment, we test video colorization on 116 video clips collected from Videvo. We apply the learning based methods for video colorization. It is too costly to use the method in~\cite{he2018deep} ($90s/\text{frame}$ compared to $0.61s/\text{frame}$ in our method), so we exclude it in this comparison. The quantitative comparison is included in Table~\ref{table:comparison_image}. We also apply the method proposed in~\cite{lai2018learning} which post-processes per-frame colorized videos and generates temporally consistent results. We denote these post-processed outputs with `$+$' in Table~\ref{table:comparison_image}. We measure the temporal stability using Eq.~\ref{eq:flow} averaged over all adjacent frame pairs in the results. A smaller temporal error represents less flickering. The post-processing method~\cite{lai2018learning} significantly reduces the temporal flickering while our method produces a comparably stable result. However, their method~\cite{lai2018learning} degrades the visual quality since the temporal filtering introduces blurriness. As shown in the example in Figure~\ref{fig:video_comparison1}, our method exhibits vibrant colors in each frame with significantly fewer artifacts compared to other methods. Meanwhile, the successively colorized frames demonstrate good temporal consistency. \vspace{-0.7em}

	\paragraph{Comparison with Color Propagation Methods.}
	In order to show that our method can degenerate to the case where the reference is a colored frame for the video itself, we compare it with two recent color propagation methods: VPN~\cite{jampani2017video} and STN~\cite{liu2018switchable}. We also include optical flow based color propagation as a baseline. Figure~\ref{fig:propagation_curve} shows the PSNR curve with frame propagation tested on the DAVIS dataset~\cite{perazzi2016benchmark}. Optical flow based methods provides the highest PSNR in the initial frames but deteriorates significantly thereafter. The methods STN and VPN also suffer from PNSR degradation. Our method with $\mathcal{L}_1$ loss attains a most stable curve, showing the capability for propagating to more frames. \vspace{-0.7em}

\paragraph{User Studies.}  We first compare our video colorization with three methods of per-frame automatic video colorization: Larsson et al.~\cite{larsson2016learning}, Zhang et al.~\cite{zhang2016colorful} and Iizuka et al.~\cite{iizuka2016let}. We used 19 videos randomly selected from the Videvo test dataset. For each video, we ask the user to rank the results generated by the four methods in terms of temporal consistency and visual photorealism. Figure~\ref{fig:user_study} (left) shows the results based on the feedback from 20 users. Our approach is $50.66\%$ more likely to be chosen as the 1st-rank result. We also compare against two video propagation methods: VPN~\cite{jampani2017video} and STN~\cite{liu2018switchable} on 15 randomly selected videos from the DAVIS test dataset. For a fair comparison, we initialize all three methods with the same colorization result of the first frame (using the ground truth). Figure~\ref{fig:user_study} (right) shows the survey results. Again, our method achieves the highest 1st-rank percentage at $79.67\%$. 



    \vspace{-.06cm}
	\section{Conclusion}
    \vspace{-.1cm}
	In this work, we propose the first exemplar-based video colorization algorithm. We unify the semantic correspondence and colorization into a single network, training it end-to-end. Our method produces temporal consistent video colorization with realistic effects. Readers could refer to our supplementary material for more quantitative results. 
    \vspace{-.2cm}
	\paragraph{Acknowledgements:}
    This work was partly supported by Hong Kong GRF Grant No. 16208814 and CityU of Hong Kong Startup Grant No. 7200607/CS. 

	{\small
		\bibliographystyle{ieeetr}
		\bibliography{colorization_paper.bbl}

\begin{thebibliography}{10}

\bibitem{levin2004colorization}
A.~Levin, D.~Lischinski, and Y.~Weiss, ``Colorization using optimization,'' in
  {\em ACM transactions on graphics (TOG)}, vol.~23, pp.~689--694, ACM, 2004.

\bibitem{yatziv2004fast}
L.~Yatziv and G.~Sapiro, ``Fast image and video colorization using chrominance
  blending,'' 2004.

\bibitem{huang2005adaptive}
Y.-C. Huang, Y.-S. Tung, J.-C. Chen, S.-W. Wang, and J.-L. Wu, ``An adaptive
  edge detection based colorization algorithm and its applications,'' in {\em
  Proceedings of the 13th annual ACM international conference on Multimedia},
  pp.~351--354, ACM, 2005.

\bibitem{qu2006manga}
Y.~Qu, T.-T. Wong, and P.-A. Heng, ``Manga colorization,'' in {\em ACM
  Transactions on Graphics (TOG)}, vol.~25, pp.~1214--1220, ACM, 2006.

\bibitem{luan2007natural}
Q.~Luan, F.~Wen, D.~Cohen-Or, L.~Liang, Y.-Q. Xu, and H.-Y. Shum, ``Natural
  image colorization,'' in {\em Proceedings of the 18th Eurographics conference
  on Rendering Techniques}, pp.~309--320, Eurographics Association, 2007.

\bibitem{welsh2002transferring}
T.~Welsh, M.~Ashikhmin, and K.~Mueller, ``Transferring color to greyscale
  images,'' in {\em ACM Transactions on Graphics (TOG)}, vol.~21, pp.~277--280,
  ACM, 2002.

\bibitem{bugeau2014variational}
A.~Bugeau, V.-T. Ta, and N.~Papadakis, ``Variational exemplar-based image
  colorization,'' {\em IEEE Transactions on Image Processing}, vol.~23, no.~1,
  pp.~298--307, 2014.

\bibitem{liu2008intrinsic}
X.~Liu, L.~Wan, Y.~Qu, T.-T. Wong, S.~Lin, C.-S. Leung, and P.-A. Heng,
  ``Intrinsic colorization,'' in {\em ACM Transactions on Graphics (TOG)},
  vol.~27, p.~152, ACM, 2008.

\bibitem{chia2011semantic}
A.~Y.-S. Chia, S.~Zhuo, R.~K. Gupta, Y.-W. Tai, S.-Y. Cho, P.~Tan, and S.~Lin,
  ``Semantic colorization with internet images,'' in {\em ACM Transactions on
  Graphics (TOG)}, vol.~30, p.~156, ACM, 2011.

\bibitem{gupta2012image}
R.~K. Gupta, A.~Y.-S. Chia, D.~Rajan, E.~S. Ng, and H.~Zhiyong, ``Image
  colorization using similar images,'' in {\em Proceedings of the 20th ACM
  international conference on Multimedia}, pp.~369--378, ACM, 2012.

\bibitem{charpiat2008automatic}
G.~Charpiat, M.~Hofmann, and B.~Sch{\"o}lkopf, ``Automatic image colorization
  via multimodal predictions,'' in {\em European conference on computer
  vision}, pp.~126--139, Springer, 2008.

\bibitem{ironi2005colorization}
R.~Ironi, D.~Cohen-Or, and D.~Lischinski, ``Colorization by example.,'' in {\em
  Rendering Techniques}, pp.~201--210, Citeseer, 2005.

\bibitem{tai2005local}
Y.-W. Tai, J.-Y. Jia, and C.-K. Tang, ``Local color transfer via probabilistic
  segmentation by expectation-maximization,'' in {\em IEEE Conference on
  Computer Vision \& Pattern Recognition (CVPR)}, 2005.

\bibitem{cheng2015deep}
Z.~Cheng, Q.~Yang, and B.~Sheng, ``Deep colorization,'' in {\em Proceedings of
  the IEEE International Conference on Computer Vision}, pp.~415--423, 2015.

\bibitem{iizuka2016let}
S.~Iizuka, E.~Simo-Serra, and H.~Ishikawa, ``Let there be color!: joint
  end-to-end learning of global and local image priors for automatic image
  colorization with simultaneous classification,'' {\em ACM Transactions on
  Graphics (TOG)}, vol.~35, no.~4, p.~110, 2016.

\bibitem{larsson2016learning}
G.~Larsson, M.~Maire, and G.~Shakhnarovich, ``Learning representations for
  automatic colorization,'' in {\em European Conference on Computer Vision},
  pp.~577--593, Springer, 2016.

\bibitem{zhang2016colorful}
R.~Zhang, P.~Isola, and A.~A. Efros, ``Colorful image colorization,'' in {\em
  European Conference on Computer Vision}, pp.~649--666, Springer, 2016.

\bibitem{deshpande2015learning}
A.~Deshpande, J.~Rock, and D.~Forsyth, ``Learning large-scale automatic image
  colorization,'' in {\em Proceedings of the IEEE International Conference on
  Computer Vision}, pp.~567--575, 2015.

\bibitem{zhao2018pixel}
J.~Zhao, L.~Liu, C.~G. Snoek, J.~Han, and L.~Shao, ``Pixel-level semantics
  guided image colorization,'' {\em arXiv preprint arXiv:1808.01597}, 2018.

\bibitem{baldassarre2017deep}
F.~Baldassarre, D.~G. Mor{\'\i}n, and L.~Rod{\'e}s-Guirao, ``Deep
  koalarization: Image colorization using cnns and inception-resnet-v2,'' {\em
  arXiv preprint arXiv:1712.03400}, 2017.

\bibitem{bonneel2015blind}
N.~Bonneel, J.~Tompkin, K.~Sunkavalli, D.~Sun, S.~Paris, and H.~Pfister,
  ``Blind video temporal consistency,'' {\em ACM Transactions on Graphics
  (TOG)}, vol.~34, no.~6, p.~196, 2015.

\bibitem{lai2018learning}
W.-S. Lai, J.-B. Huang, O.~Wang, E.~Shechtman, E.~Yumer, and M.-H. Yang,
  ``Learning blind video temporal consistency,'' {\em arXiv preprint
  arXiv:1808.00449}, 2018.

\bibitem{sheng2014video}
B.~Sheng, H.~Sun, M.~Magnor, and P.~Li, ``Video colorization using parallel
  optimization in feature space,'' {\em IEEE Transactions on Circuits and
  Systems for Video Technology}, vol.~24, no.~3, pp.~407--417, 2014.

\bibitem{dougan2015key}
P.~Do{\u{g}}an, T.~O. Ayd{\i}n, N.~Stefanoski, and A.~Smolic, ``Key-frame based
  spatiotemporal scribble propagation,'' in {\em Proceedings of the
  Eurographics Workshop on Intelligent Cinematography and Editing}, pp.~13--20,
  Eurographics Association, 2015.

\bibitem{paul2017spatiotemporal}
S.~Paul, S.~Bhattacharya, and S.~Gupta, ``Spatiotemporal colorization of video
  using 3d steerable pyramids,'' {\em IEEE Transactions on Circuits and Systems
  for Video Technology}, vol.~27, no.~8, pp.~1605--1619, 2017.

\bibitem{jampani2017video}
V.~Jampani, R.~Gadde, and P.~V. Gehler, ``Video propagation networks,'' in {\em
  Proc. CVPR}, vol.~6, p.~7, 2017.

\bibitem{vondrick2018tracking}
C.~Vondrick, A.~Shrivastava, A.~Fathi, S.~Guadarrama, and K.~Murphy, ``Tracking
  emerges by colorizing videos,'' in {\em Proc. ECCV}, 2018.

\bibitem{liu2018switchable}
S.~Liu, G.~Zhong, S.~De~Mello, J.~Gu, V.~Jampani, M.-H. Yang, and J.~Kautz,
  ``Switchable temporal propagation network,'' {\em arXiv preprint
  arXiv:1804.08758}, 2018.

\bibitem{meyer2018deep}
S.~Meyer, V.~Cornill{\`e}re, A.~Djelouah, C.~Schroers, and M.~Gross, ``Deep
  video color propagation,'' {\em arXiv preprint arXiv:1808.03232}, 2018.

\bibitem{he2018deep}
M.~He, D.~Chen, J.~Liao, P.~V. Sander, and L.~Yuan, ``Deep exemplar-based
  colorization,'' {\em ACM Transactions on Graphics (TOG)}, vol.~37, no.~4,
  p.~47, 2018.

\bibitem{liao2017visual}
J.~Liao, Y.~Yao, L.~Yuan, G.~Hua, and S.~B. Kang, ``Visual attribute transfer
  through deep image analogy,'' {\em arXiv preprint arXiv:1705.01088}, 2017.

\bibitem{zhang2017real}
R.~Zhang, J.-Y. Zhu, P.~Isola, X.~Geng, A.~S. Lin, T.~Yu, and A.~A. Efros,
  ``Real-time user-guided image colorization with learned deep priors,'' {\em
  arXiv preprint arXiv:1705.02999}, 2017.

\bibitem{he2017neural}
M.~He, J.~Liao, L.~Yuan, and P.~V. Sander, ``Neural color transfer between
  images,'' {\em arXiv preprint arXiv:1710.00756}, 2017.

\bibitem{isola2017image}
P.~Isola, J.-Y. Zhu, T.~Zhou, and A.~A. Efros, ``Image-to-image translation
  with conditional adversarial networks,'' {\em arXiv preprint}, 2017.

\bibitem{deshpande2017learning}
A.~Deshpande, J.~Lu, M.-C. Yeh, M.~J. Chong, and D.~A. Forsyth, ``Learning
  diverse image colorization.,'' in {\em CVPR}, pp.~2877--2885, 2017.

\bibitem{messaoud2018structural}
S.~Messaoud, D.~Forsyth, and A.~G. Schwing, ``Structural consistency and
  controllability for diverse colorization,'' {\em arXiv preprint
  arXiv:1809.02129}, 2018.

\bibitem{guadarrama2017pixcolor}
S.~Guadarrama, R.~Dahl, D.~Bieber, M.~Norouzi, J.~Shlens, and K.~Murphy,
  ``Pixcolor: Pixel recursive colorization,'' {\em arXiv preprint
  arXiv:1705.07208}, 2017.

\bibitem{royer2017probabilistic}
A.~Royer, A.~Kolesnikov, and C.~H. Lampert, ``Probabilistic image
  colorization,'' {\em arXiv preprint arXiv:1705.04258}, 2017.

\bibitem{jacob2009colorization}
V.~G. Jacob and S.~Gupta, ``Colorization of grayscale images and videos using a
  semiautomatic approach,'' in {\em Image Processing (ICIP), 2009 16th IEEE
  International Conference on}, pp.~1653--1656, IEEE, 2009.

\bibitem{ben2015approximate}
N.~Ben-Zrihem and L.~Zelnik-Manor, ``Approximate nearest neighbor fields in
  video,'' in {\em Proceedings of the IEEE Conference on Computer Vision and
  Pattern Recognition}, pp.~5233--5242, 2015.

\bibitem{xia2016robust}
S.~Xia, J.~Liu, Y.~Fang, W.~Yang, and Z.~Guo, ``Robust and automatic video
  colorization via multiframe reordering refinement,'' in {\em Image Processing
  (ICIP), 2016 IEEE International Conference on}, pp.~4017--4021, IEEE, 2016.

\bibitem{simonyan2014very}
K.~Simonyan and A.~Zisserman, ``Very deep convolutional networks for
  large-scale image recognition,'' {\em arXiv preprint arXiv:1409.1556}, 2014.

\bibitem{wang2017non}
X.~Wang, R.~Girshick, A.~Gupta, and K.~He, ``Non-local neural networks,'' {\em
  arXiv preprint arXiv:1711.07971}, vol.~10, 2017.

\bibitem{johnson2016perceptual}
J.~Johnson, A.~Alahi, and L.~Fei-Fei, ``Perceptual losses for real-time style
  transfer and super-resolution,'' in {\em European Conference on Computer
  Vision}, pp.~694--711, Springer, 2016.

\bibitem{mechrez2018contextual}
R.~Mechrez, I.~Talmi, and L.~Zelnik-Manor, ``The contextual loss for image
  transformation with non-aligned data,'' {\em arXiv preprint
  arXiv:1803.02077}, 2018.

\bibitem{farbman2008edge}
Z.~Farbman, R.~Fattal, D.~Lischinski, and R.~Szeliski, ``Edge-preserving
  decompositions for multi-scale tone and detail manipulation,'' in {\em ACM
  Transactions on Graphics (TOG)}, vol.~27, p.~67, ACM, 2008.

\bibitem{jolicoeur2018relativistic}
A.~Jolicoeur-Martineau, ``The relativistic discriminator: a key element missing
  from standard gan,'' {\em arXiv preprint arXiv:1807.00734}, 2018.

\bibitem{chen2017coherent}
D.~Chen, J.~Liao, L.~Yuan, N.~Yu, and G.~Hua, ``Coherent online video style
  transfer,'' in {\em Proceedings of the IEEE International Conference on
  Computer Vision}, pp.~1105--1114, 2017.

\bibitem{zhang2018self}
H.~Zhang, I.~Goodfellow, D.~Metaxas, and A.~Odena, ``Self-attention generative
  adversarial networks,'' {\em arXiv preprint arXiv:1805.08318}, 2018.

\bibitem{miyato2018spectral}
T.~Miyato, T.~Kataoka, M.~Koyama, and Y.~Yoshida, ``Spectral normalization for
  generative adversarial networks,'' {\em arXiv preprint arXiv:1802.05957},
  2018.

\bibitem{Videvo}
``Videvo.'' \url{https://www.videvo.net/}.

\bibitem{marszalek09}
M.~Marsza{\l}ek, I.~Laptev, and C.~Schmid, ``Actions in context,'' in {\em IEEE
  Conference on Computer Vision \& Pattern Recognition}, 2009.

\bibitem{ilg2017flownet}
E.~Ilg, N.~Mayer, T.~Saikia, M.~Keuper, A.~Dosovitskiy, and T.~Brox, ``Flownet
  2.0: Evolution of optical flow estimation with deep networks,'' in {\em IEEE
  conference on computer vision and pattern recognition (CVPR)}, vol.~2, p.~6,
  2017.

\bibitem{ruder2016artistic}
M.~Ruder, A.~Dosovitskiy, and T.~Brox, ``Artistic style transfer for videos,''
  in {\em German Conference on Pattern Recognition}, pp.~26--36, Springer,
  2016.

\bibitem{heusel2017gans}
M.~Heusel, H.~Ramsauer, T.~Unterthiner, B.~Nessler, and S.~Hochreiter, ``Gans
  trained by a two time-scale update rule converge to a local nash
  equilibrium,'' in {\em Advances in Neural Information Processing Systems},
  pp.~6626--6637, 2017.

\bibitem{hasler2003measuring}
D.~Hasler and S.~E. Suesstrunk, ``Measuring colorfulness in natural images,''
  in {\em Human vision and electronic imaging VIII}, vol.~5007, pp.~87--96,
  International Society for Optics and Photonics, 2003.

\bibitem{perazzi2016benchmark}
F.~Perazzi, J.~Pont-Tuset, B.~McWilliams, L.~Van~Gool, M.~Gross, and
  A.~Sorkine-Hornung, ``A benchmark dataset and evaluation methodology for
  video object segmentation,'' in {\em Proceedings of the IEEE Conference on
  Computer Vision and Pattern Recognition}, pp.~724--732, 2016.

\end{thebibliography}
	}

	\newpage
	\onecolumn

	\begin{appendices}
	\section{Details of network architecture}
	The overall network consists of two sub-modules: the correspondence subnet and the colorization subnet. The correspondence network receives the VGG features for both input image and the reference image. The multi-level features are fused through multiple residual blocks for the two images, then the warped reference color and the similarity map can be calculated from the correlation matrix using the features from two images. We use reflective padding to reduce the boundary artifact and employ PReLU as activation function within the residual blocks. Instance normalization is used after each convolutional layer. The architecture of the correspondence subnet is detailed in Table~\ref{table:correspondence_net}.

\begin{table*}[!htb]
\small
\centering
\begin{tabular}{cccccccc}
\hline\hline
\multicolumn{4}{c|}{input image}                                                                                                             & \multicolumn{4}{c}{reference image}                                                                                    \\ \hline
\multicolumn{1}{c|}{relu2-2}        & \multicolumn{1}{c|}{relu3-2}   & \multicolumn{1}{c|}{relu4-2}      & \multicolumn{1}{c|}{relu5-2}      & \multicolumn{1}{c|}{relu2-2}        & \multicolumn{1}{c|}{relu3-2}   & \multicolumn{1}{c|}{relu4-2}      & relu5-2      \\ \hline
\multicolumn{1}{c|}{conv, 128}      & \multicolumn{1}{c|}{conv, 128} & \multicolumn{1}{c|}{conv, 256}    & \multicolumn{1}{c|}{conv up, 256} & \multicolumn{1}{c|}{conv, 128}      & \multicolumn{1}{c|}{conv, 128} & \multicolumn{1}{c|}{conv, 256}    & conv up, 256 \\ \hline
\multicolumn{1}{c|}{conv, stride 2, 256} & \multicolumn{1}{c|}{conv, 256} & \multicolumn{1}{c|}{conv up, 256} & \multicolumn{1}{c|}{conv up, 256} & \multicolumn{1}{c|}{conv, stride 2, 256} & \multicolumn{1}{c|}{conv, 256} & \multicolumn{1}{c|}{conv up, 256} & conv up, 256 \\ \hline
\multicolumn{4}{c|}{concatenate}                                                                                                             & \multicolumn{4}{c}{concatenate}                                                                                        \\ \hline
\multicolumn{4}{c|}{Resblock, 256}                                                                                                           & \multicolumn{4}{c}{Resblock, 256}                                                                                      \\ \hline
\multicolumn{4}{c|}{Resblock, 256}                                                                                                           & \multicolumn{4}{c}{Resblock, 256}                                                                                      \\ \hline
\multicolumn{4}{c|}{Resblock, 256}                                                                                                           & \multicolumn{4}{c}{Resblock, 256}                                                                                      \\ \hline
\multicolumn{8}{c}{correlation matrix}                                                                                                                                                                                                                                 \\ \hline
\multicolumn{8}{c}{softmax}                                                                                                                                                                                                                                            \\ \hline
\multicolumn{8}{c}{color warping \& similarity map}                                                                                                                                                                                                                    \\ \hline\hline
\end{tabular}
\label{table:correspondence_net}
\vspace{0.5em}
\caption{The architecture of correspondence subnet.}
\end{table*}

	The colorization subnet receives four inputs - the luminance of the current frame $x_t^l$, the warped reference color $\mathcal{W}^{ab}$, the similarity map $\mathcal{S}$ and the colorized last frame $\tilde{x}_{t-1}^{ab}$. We adopt an architecture similar as~\cite{zhang2017real}, which is an auto-encoder with skip-connections to reuse the low-level features. The encoder hierarchically captures the multi-level features and the decode predicts the chrominance of the current frame by utilizing the multi-level features. The network consists of several convolutional block, each of them containing 2$\sim$3 convolutional layers. Still, we add instance normalize after each convolutional block. At the bottleneck, we also introduce local skip connections in addition to the global skip connections between the encoder and decoder, which helps ease the gradient flow within the network. The structure details are shown in Table~\ref{table:colorization_net}.
	
	We train a discriminator network in an adversarial manner. The structure of the discriminator is shown in Table~\ref{table:discriminator}. To take the global information into account, we insert the self-attention layer at the second layer of the discriminator. At the start of the training, we first exclude the adversarial loss since the colorized images can be easily identified to be fake. After training several epochs, we include the adversarial loss and the colorized result become more vivid and realistic.

\begin{table*}[!htb]
\centering
\small
\begin{tabular}{c|c}
\hline\hline
\multicolumn{2}{c} {current frame $x_t$, warped color $\mathcal{W}^{ab}$, similarity map $\mathcal{S}$ and colorized last frame $x_{t-1}^{ab}$} \\ \hline
conv block1     & conv*2, channel=64, downscale 1/2                                  \\ \hline
conv block2     & conv*2, channel=128, downscale 1/2                                 \\ \hline
conv block3     & conv*3, channel=256, downscale 1/2                                 \\ \hline
Residual block1 & conv*3, channel=512                                                \\ \hline
Residual block2 & conv*3, channel=512                                                \\ \hline
Residual block3 & conv*3, channel=512                                                \\ \hline
conv block 7    & skip connection from conv block 3, conv*3, channel=256, upscale x2 \\ \hline
conv block 8    & skip connection from conv block 2, conv*2, channel=128, upscale x2 \\ \hline
conv block 9    & skip connection from conv block 1, conv*2, channel=64, upscale x2  \\ \hline
conv            & conv*1, channel=2                                                  \\ \hline
tanh            &                                                                    \\ \hline\hline
\end{tabular}%
\vspace{0.5em}
\caption{The architecture of the colorization subnet.}
\label{table:colorization_net}
\end{table*}
	
\begin{table}[]
\centering
\begin{tabular}{c}
\hline\hline
video frames \{$\tilde{x}_{t},\tilde{x}_{t-1}$\} \\ \hline
conv, kernel=4, stride=2, channel=64                       \\ \hline
self-attention                                   \\ \hline
conv, kernel=4, stride=2, channel=64                       \\ \hline
conv, kernel=4, stride=2, channel=128                      \\ \hline
conv, kernel=4, stride=2, channel=256                      \\ \hline
conv, kernel=4, stride=2, channel=512                      \\ \hline
conv, kernel=4, stride=2, channel=1024                     \\ \hline
conv, kernel=3, channel=1                                  \\ \hline
average pooling                                  \\ \hline\hline
\end{tabular}
\vspace{0.5em}
\caption{The discriminator architecture}
\label{table:discriminator}
\end{table}
	
	\section{Multimodal colorization}
\begin{figure*}[!tbh]
    \setlength\tabcolsep{1.5pt}
    \centering
    \small
    \begin{tabular}{ccccc}
    ground truth & reference & result & reference & result  \\
    \includegraphics[width=0.18\columnwidth]{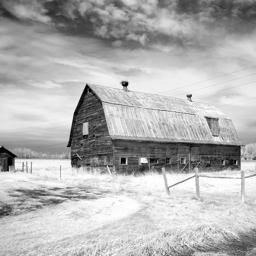}&
    \includegraphics[width=0.18\columnwidth]{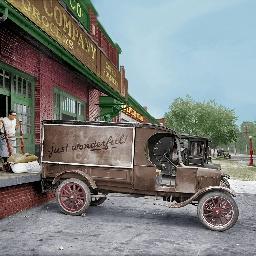}&
    \includegraphics[width=0.18\columnwidth]{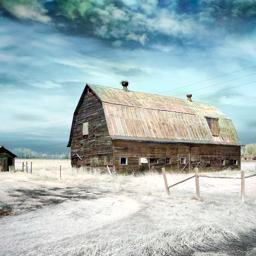}&
    \includegraphics[width=0.18\columnwidth]{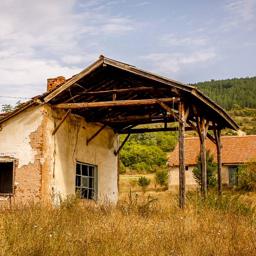}&
    \includegraphics[width=0.18\columnwidth]{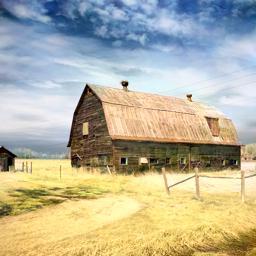}\\
    &
    \includegraphics[width=0.18\columnwidth]{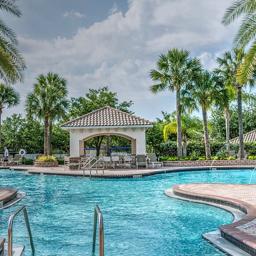}&
    \includegraphics[width=0.18\columnwidth]{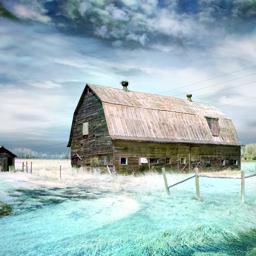}&
    \includegraphics[width=0.18\columnwidth]{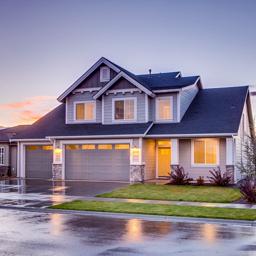}&
    \includegraphics[width=0.18\columnwidth]{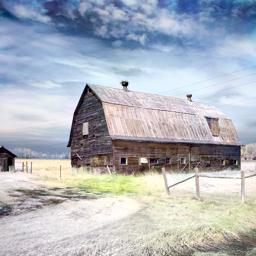}\\
    \end{tabular}
    \caption{Multi-modal colorization according to the user reference.}
    \label{fig:multimodal}
  	\end{figure*}

\begin{figure*}[!tbh]
    \setlength\tabcolsep{1.5pt}
    \centering
    \small
    \begin{tabular}{ccccc}
    ground truth & reference & result & reference & result  \\
    \includegraphics[width=0.18\columnwidth]{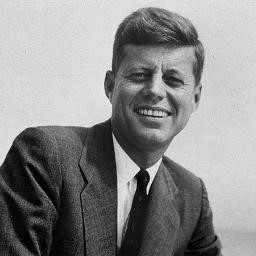}&
    \includegraphics[width=0.18\columnwidth]{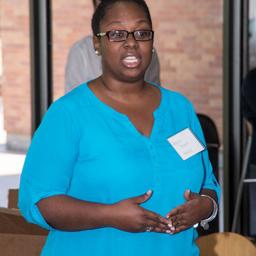}&
    \includegraphics[width=0.18\columnwidth]{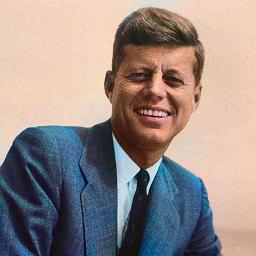}&
    \includegraphics[width=0.18\columnwidth]{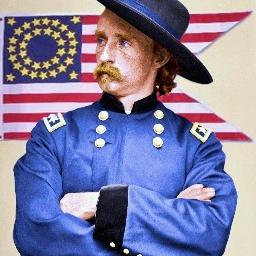}&
    \includegraphics[width=0.18\columnwidth]{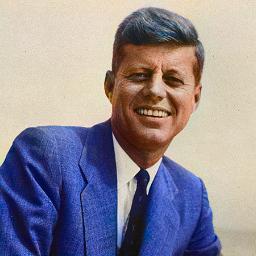}\\
    &
    \includegraphics[width=0.18\columnwidth]{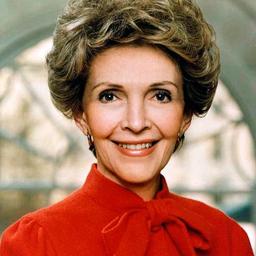}&
    \includegraphics[width=0.18\columnwidth]{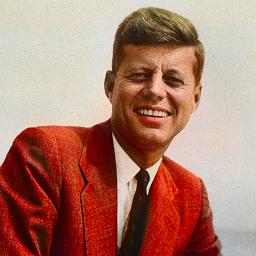}&
    \includegraphics[width=0.18\columnwidth]{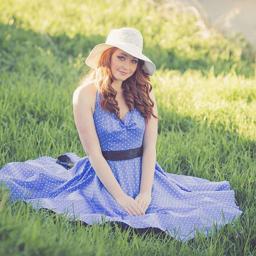}&
    \includegraphics[width=0.18\columnwidth]{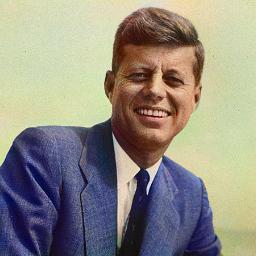}\\
    \end{tabular}
    \caption{Multi-modal colorization according to the user reference.}
    \label{fig:multimodal1}
  	\end{figure*}

\begin{figure*}[!tbh]
    \setlength\tabcolsep{1.5pt}
    \centering
    \small
    \begin{tabular}{ccccc}
    ground truth & reference & result & reference & result  \\
    \includegraphics[width=0.18\columnwidth]{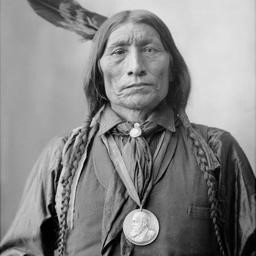}&
    \includegraphics[width=0.18\columnwidth]{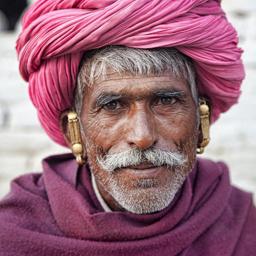}&
    \includegraphics[width=0.18\columnwidth]{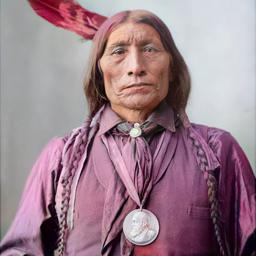}&
    \includegraphics[width=0.18\columnwidth]{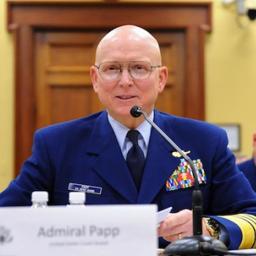}&
    \includegraphics[width=0.18\columnwidth]{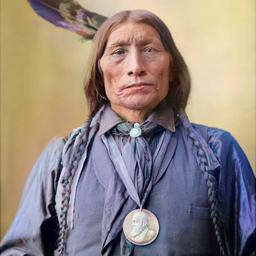}\\
    &
    \includegraphics[width=0.18\columnwidth]{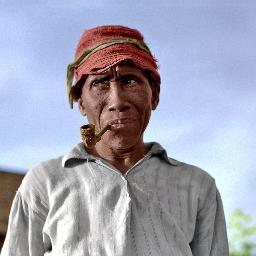}&
    \includegraphics[width=0.18\columnwidth]{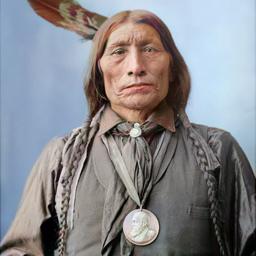}&
    \includegraphics[width=0.18\columnwidth]{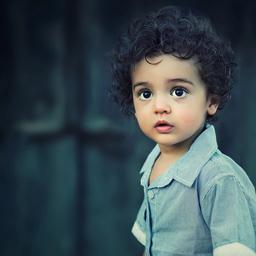}&
    \includegraphics[width=0.18\columnwidth]{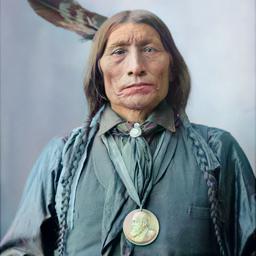}\\
    \end{tabular}
    \caption{Multi-modal colorization according to the user reference.}
    \label{fig:multimodal2}
  	\end{figure*}

\begin{figure*}[!tbh]
    \setlength\tabcolsep{1.5pt}
    \centering
    \small
    \begin{tabular}{ccccc}
    ground truth & reference & result & reference & result  \\
    \includegraphics[width=0.18\columnwidth]{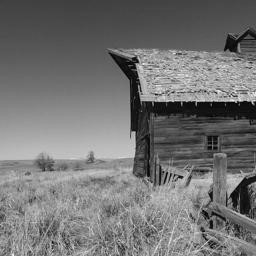}&
    \includegraphics[width=0.18\columnwidth]{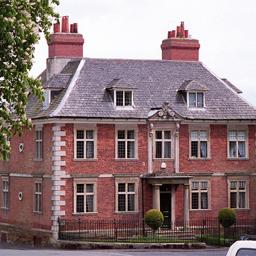}&
    \includegraphics[width=0.18\columnwidth]{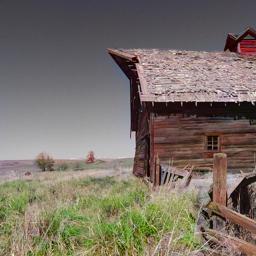}&
    \includegraphics[width=0.18\columnwidth]{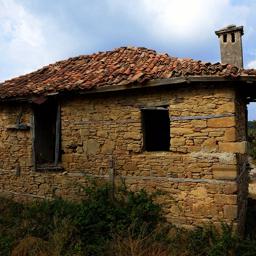}&
    \includegraphics[width=0.18\columnwidth]{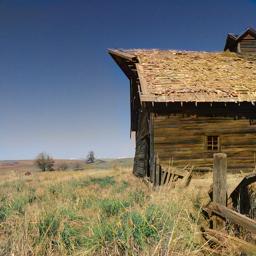}\\
    &
    \includegraphics[width=0.18\columnwidth]{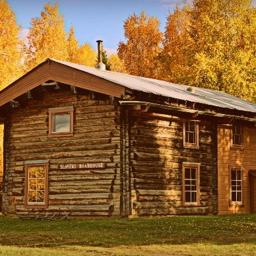}&
    \includegraphics[width=0.18\columnwidth]{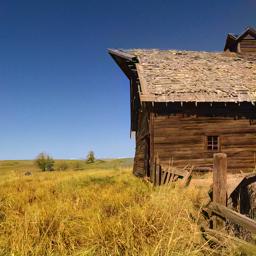}&
    \includegraphics[width=0.18\columnwidth]{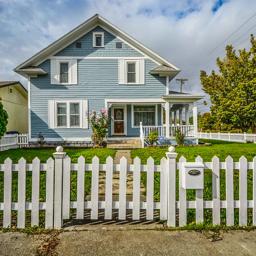}&
    \includegraphics[width=0.18\columnwidth]{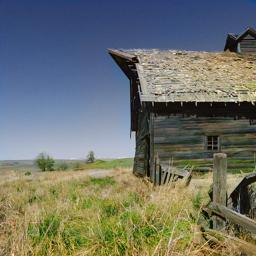}\\
    \end{tabular}
    \caption{Multi-modal colorization according to the user reference.}
    \label{fig:multimodal3}
	\end{figure*}

	Our method allows users to customize the colorization result by providing a reference. Figure~\ref{fig:multimodal}-\ref{fig:multimodal3} shows more examples. Our methods demonstrates realistic quality by properly propagating the colors from the corresponding regions in the reference.
  	
\section{Colorization on legacy video}
\begin{figure*}[!tbh]
    \setlength\tabcolsep{1.5pt}
    \centering
    \small
    \begin{tabular}{ccc}
    legacy video frame & reference & output  \\
    \includegraphics[height=1.2in]{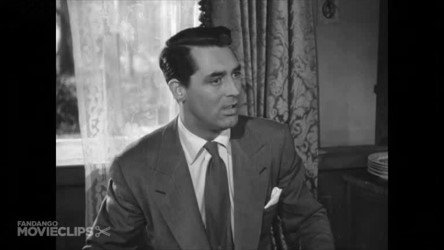}&
    \includegraphics[height=1.2in]{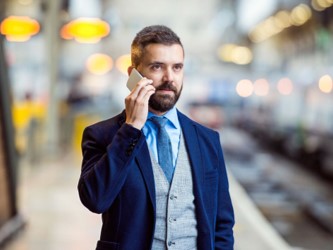}&
	\includegraphics[height=1.2in]{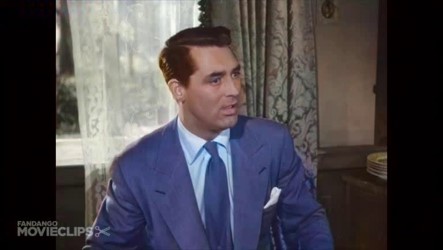}\\
	\includegraphics[height=1.2in]{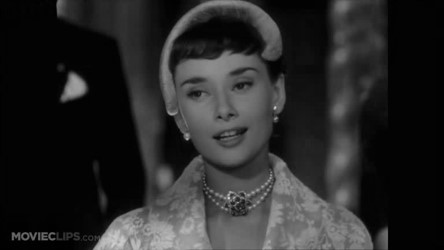}&
    \includegraphics[height=1.2in]{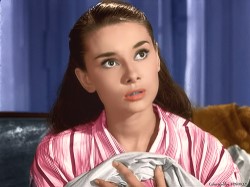}&
    \includegraphics[height=1.2in]{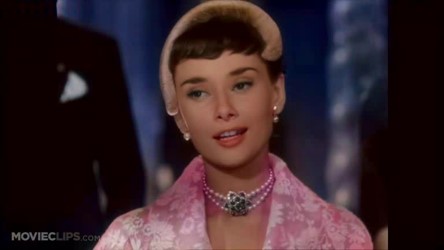}\\
    \end{tabular}
    \caption{Colorization on legacy videos.}
    \label{fig:legacy}
	\end{figure*}

Though our model is trained on imagenet and downloaded vidoes, it can colorize real legacy videos with impressive quality as well. As shown in Fig.~\ref{fig:legacy}, our method revives the legacy videos with vivid color faithfully to the given reference. 

\section{Quantitative ablation study}
The quantitative results are provided in Table~\ref{fig:quantitative}. The full model achieves the best perceptual quality with the lowest FID score. $\mathcal{L}_{adv}$ greatly improves the saturation though it sacrifices the recognition accuracy by a small amount. Note that the effects of $\mathcal{L}_{perc}$ and $\mathcal{L}_{context}$ are contradictory ($\mathcal{L}_{perc}\rightarrow$ input semantics, $\mathcal{L}_{context}\rightarrow$ reference style), so the result using the full model is slightly less saturated.

\begin{table}[!tb]
\centering
\small
\setlength{\tabcolsep}{3pt}
\caption{Quantitative result for ablation study.}
\begin{tabular}{lcccc}
\toprule
& Top-5 Acc(\%) &  Top-1 Acc(\%) & FID & Colorful \\
\cline{1-5}
w/o $\mathcal{L}_{smooth}$  & 85.28  & 64.48  & 4.12 & 17.12\\
w/o $\mathcal{L}_{perc}$  & 83.96  &  63.29 & 5.23 & 18.44\\
w/o $\mathcal{L}_{context}$  & 84.67 & 62.96 & 6.06 & 18.94 \\
w/o $\mathcal{L}_{adv}$  & 86.30 & 65.46 & 4.33 & 14.93 \\
Full  & 85.82 & 64.64 & 4.02 & 17.90 \\
\bottomrule
\label{fig:quantitative}
\end{tabular}
\end{table}

	\section{User study}
	
		We conduct two user studies: one to measure the video colorization quality and another for video propagation. For the first study, we first compare our video colorization with three methods of per-frame automatic video colorization: Larsson et al.~\cite{larsson2016learning}, Zhang et al.~\cite{zhang2016colorful} and Iizuka et al.~\cite{iizuka2016let}. We use 19 videos randomly selected from the video test dataset. For each video, we ask the user to rank the results generated by these four methods in terms of temporal consistency and visual photorealism. Table~\ref{table:user_study1} shows details of the user study result on this task. Our approach is $50.66\%$ more likely to be chosen as the 1st-rank result which significantly outperforms all three in terms of average rank: $1.98\pm1.36$ vs. $2.39\pm1.03$ for~\cite{larsson2016learning}, $2.68\pm0.93$ for~\cite{iizuka2016let}, and $2.95\pm1.16$ for~\cite{zhang2016colorful}.

    In the second study, we compare against two video propagation methods:  VPN~\cite{jampani2017video} and STN~\cite{liu2018switchable} on 15 randomly selected videos from the DAVIS test dataset. For a fair comparison, we initialize all three methods with the same colorization result of the first frame (using the ground truth video). Users are then asked to rank these results with the same evaluation criteria as in the first study. Table~\ref{table:user_study2} shows these results.
    Again, our method achieves the highest 1st-rank percentage at $79.67\%$. Overall, it achieves the highest average rank: $1.24\pm0.156$  vs. $2.07\pm0.33$ for STN, and $2.69\pm0.36$ for VPN.
    
    In summary, users prefer the quality of our video colorization methods than other state-of-the-art methods in both cases.

	\begin{table}[!htb]
    \setlength\tabcolsep{1.5pt}
    \centering
    \small
    \begin{tabular}{c|c|c|c|c}
    \hline
    & Larsson et al.& Iizuka et al. & Zhang et al.& Ours\\
    \hline
    Top1&0.152&0.13&0.14&0.51\\
    Top2&0.34&0.159&0.18&0.19\\
    Top3&0.156&0.36&0.156&0.12\\
    Top4&0.17&0.153&0.42&0.18\\
    \hline
    Avg. rank&2.39$\pm$1.03&2.68$\pm$0.93&2.95$\pm$1.16&1.98$\pm$1.36\\
    \hline
    \end{tabular}
    \vspace{0.5em}
    \caption{User study result on comparison with the automatic video colorization}
    \label{table:user_study1}
    \end{table}
    
    \begin{table}[!htb]
    \setlength\tabcolsep{1.5pt}
    \centering
    \small
    \begin{tabular}{c|c|c|c}
    \hline
    & STN & VPN & Ours\\
    \hline
    Top1&0.13&0.07&0.8\\
    Top2&0.66&0.17&0.17\\
    Top3&0.151&0.76&0.05\\
    \hline
    Avg. rank &2.07$\pm$0.33&2.69$\pm$0.36&1.24$\pm$0.156\\
    \hline
    \end{tabular}
    \vspace{0.5em}
    \caption{User study result on comparison with the video color propagation}
    \label{table:user_study2}
    \end{table}
    

    \section{Failure case}
\begin{figure*}[!tbh]
    \setlength\tabcolsep{1.5pt}
    \centering
    \small
    \begin{tabular}{cccc}
    \includegraphics[width=0.24\columnwidth]{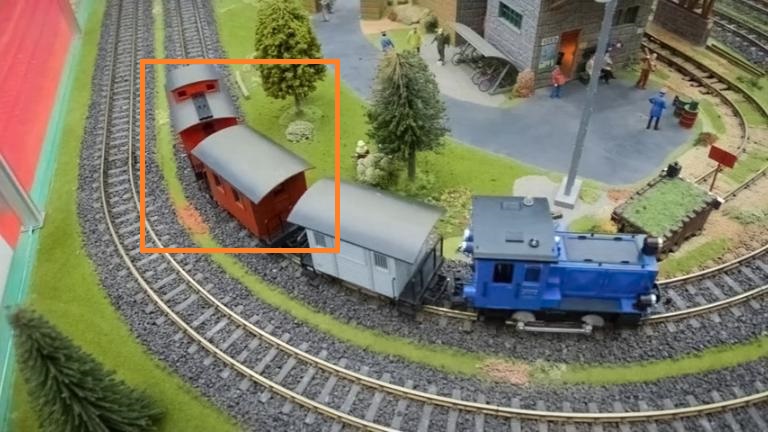}&
    \includegraphics[width=0.24\columnwidth]{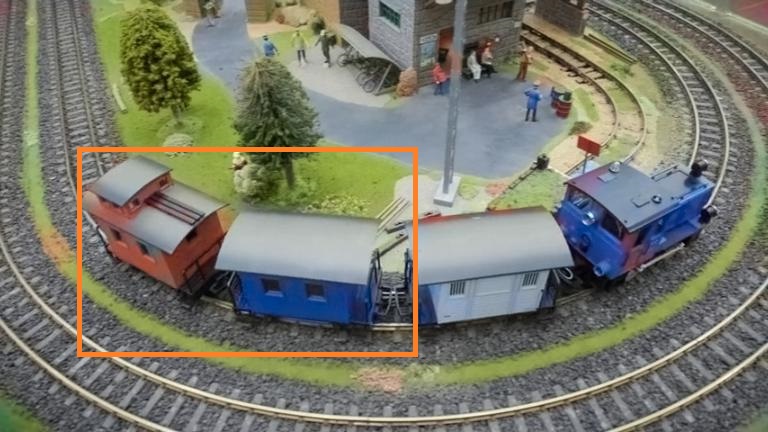}&
    \includegraphics[width=0.24\columnwidth]{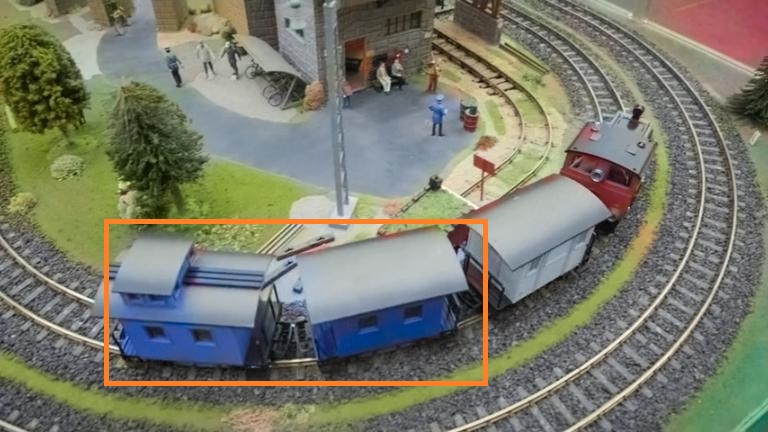}&
    \includegraphics[width=0.24\columnwidth]{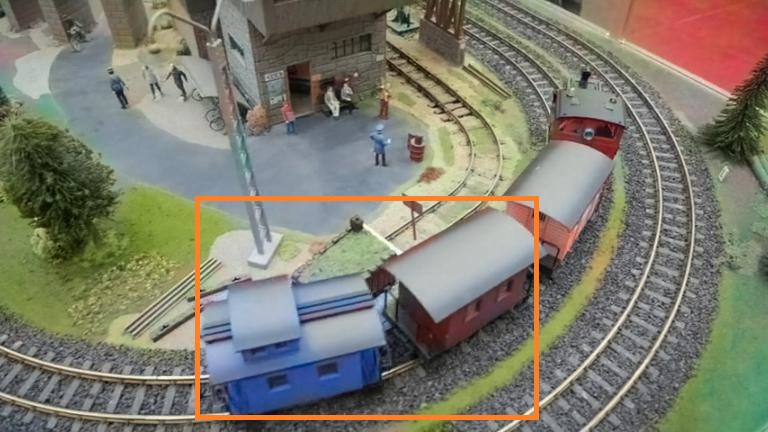}
    \end{tabular}
    \caption{Limitation: our method cannot assure long-term temporal consistency. The color of the train gradually changes (from red to blue and back to red) as the matching error accumulates.}
    \label{fig:limitation}
	  \end{figure*}

    During the training we constrain the temporal consistency of adjacent two frames. However, our method may suffer from long-term temporal inconsistency. As shown in Figure~\ref{fig:limitation} the object color gradually changes due to the mismatch of the correspondence. This is because we find the dense correspondence frame by frame and only consider the temporal effect within the colorization subnet. To deal with this issue, the temporal consistency must be incorporated as well when finding the dense correspondence. We leave this work for future exploration.
    
	\end{appendices}	
	\end{document}